\documentclass[]{fairmeta}
\usepackage{microtype}
\usepackage{graphicx}
\usepackage{booktabs} 

\usepackage{algorithm}
\usepackage{algorithmic}

\usepackage{amsmath}
\usepackage{amssymb}
\usepackage{mathtools}
\usepackage{amsthm}
\usepackage{xspace}
\usepackage{wrapfig}
\usepackage{multirow}
\usepackage{makecell}
\usepackage{lscape}

\usepackage[textsize=tiny]{todonotes}

\usepackage{pifont}
\newcommand{\cmark}{\ding{51}}
\newcommand{\xmark}{\ding{55}}

\usepackage{amsmath,amsfonts,bm}

\def\1{\bm{1}}

\def\eps{{\epsilon}}

\DeclareMathAlphabet{\mathsfit}{\encodingdefault}{\sfdefault}{m}{sl}
\SetMathAlphabet{\mathsfit}{bold}{\encodingdefault}{\sfdefault}{bx}{n}

\newcommand{\E}{\mathbb{E}}

\DeclareMathOperator*{\argmax}{arg\,max}
\DeclareMathOperator*{\argmin}{arg\,min}

\newcommand{\deq}{\mathrel{\mathop{:}}=}

\newcommand{\calB}{\mathcal{B}}

\definecolor{bostonuniversityred}{rgb}{0.8, 0.0, 0.0}

\definecolor{ao}{rgb}{0.0, 0.5, 0.0}

\definecolor{bondiblue}{rgb}{0.0, 0.6235294117647059, 0.6941176470588235}
\definecolor{pantonegreen}{rgb}{0.0, 0.6627450980392157, 0.2784313725490196}

\definecolor{oceanblueboat}{rgb}{0.0, 0.4823529411764706, 0.7686274509803922}

\newcommand{\implacro}{ASAC\xspace}
\usepackage{adjustbox}

\newcommand{\benchname}{Multi Objective Offline DMC\xspace} 
\newcommand{\benchacro}{MOOD\xspace}

\newcommand*{\rom}[1]{\expandafter\@slowromancap\romannumeral #1@}

\usepackage[toc,page,header]{appendix}
\usepackage{minitoc}

\title{Simple Ingredients for Offline Reinforcement Learning}

\author[2,*]{Edoardo Cetin}
\author[1]{Andrea Tirinzoni}
\author[1]{Matteo Pirotta}
\author[1]{Alessandro Lazaric}
\author[1,\dagger]{Yann Ollivier}
\author[1,\dagger]{Ahmed Touati}

\affiliation[1]{FAIR at Meta}
\affiliation[2]{King’s College London}

\contribution[*]{Work done at Meta}
\contribution[\dagger]{Joint last author}

\abstract{
    Offline reinforcement learning algorithms have
    proven effective on datasets highly connected to
    the target downstream task. Yet, leveraging a
    novel testbed (MOOD) in which trajectories come
    from heterogeneous sources, we show that existing methods struggle with diverse data: their
    performance considerably \textit{deteriorates} as data
    collected for related but different tasks is simply
    \textit{added} to the offline buffer. In light of this finding, we conduct a large empirical study where we
    formulate and test several hypotheses to explain
    this failure. Surprisingly, we find that scale, more
    than algorithmic considerations, is the key factor
    influencing performance. We show that simple
    methods like AWAC and IQL with increased network size overcome the paradoxical failure modes
    from the inclusion of additional data in MOOD,
    and notably outperform prior state-of-the-art algorithms on the canonical D4RL benchmark.
}

\begin{document}
\doparttoc 
\faketableofcontents 

\maketitle


\section{Introduction}

\label{sec1:intro}

\begin{figure}[h]
  \begin{minipage}{0.55\textwidth}
    Offline reinforcement learning (RL) holds the promise of overcoming the
costs and dangers of direct interaction with the environment by training agents exclusively on logged data.
However, naively applying off-policy algorithms to this setting has been shown prone to instabilities due to their natural tendency to
extrapolate beyond the given data \citep{bcq, bear}. To address this issue, \emph{policy constrained} methods propose minimal modifications to off-policy actor-critic algorithms aimed at keeping the learned policy close to the
data distribution~\citep{offline_rl_review,td3_bc, awac, IQL, XQL,td7}. 

  \end{minipage}
  \hfill
  \begin{minipage}{0.4\textwidth}
    \centering
  \includegraphics[width=0.6\textwidth]{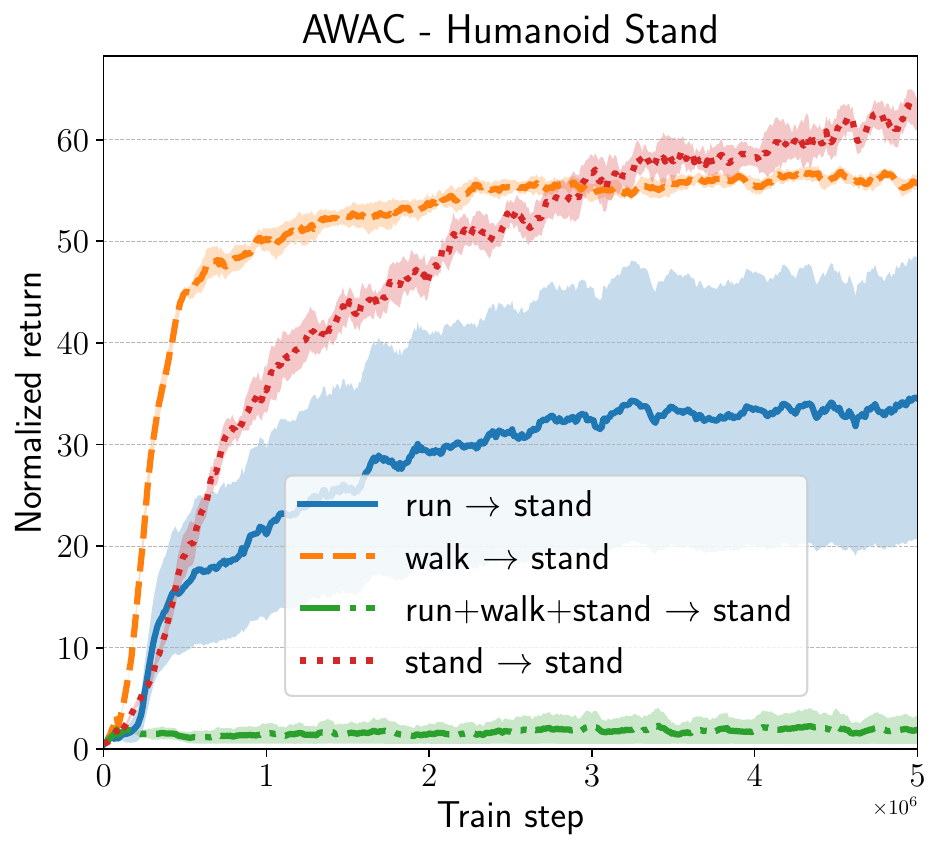}
  \caption{The AWAC algorithm learns to stand when trained on data generated by an agent learning to either stand, walk, or run, but completely fails on the union of these three datasets.}
  \label{sec3:fig:hum_stand_cross}
\end{minipage}
\end{figure}
   
For instance, TD3+Behavior Cloning
\citep[TD3+BC,][]{td3_bc} achieves this by regularizing the actor loss
with the divergence between the learned policy and the data-generating
policy, while Advantage Weighted Actor Critic \citep[AWAC,][]{awac} seeks
a policy maximizing the data likelihood weighted by its exponentiated advantage
function. Later extensions of AWAC also modify the critic loss to avoid querying
actions outside the given data by learning a value function, e.g., by expectile regression in Implicit Q-learning \citep[IQL,][]{IQL} and Gumbel regression in Extreme Q-learning \citep[XQL,][]{XQL}. 
This class of methods can be easily integrated with online fine-tuning, even leading to several successful applications for real-world tasks \citep{aw-opt, iql_industrial_ins}.

\begin{figure*}[t]
  \centering
\includegraphics[width=0.98\linewidth]{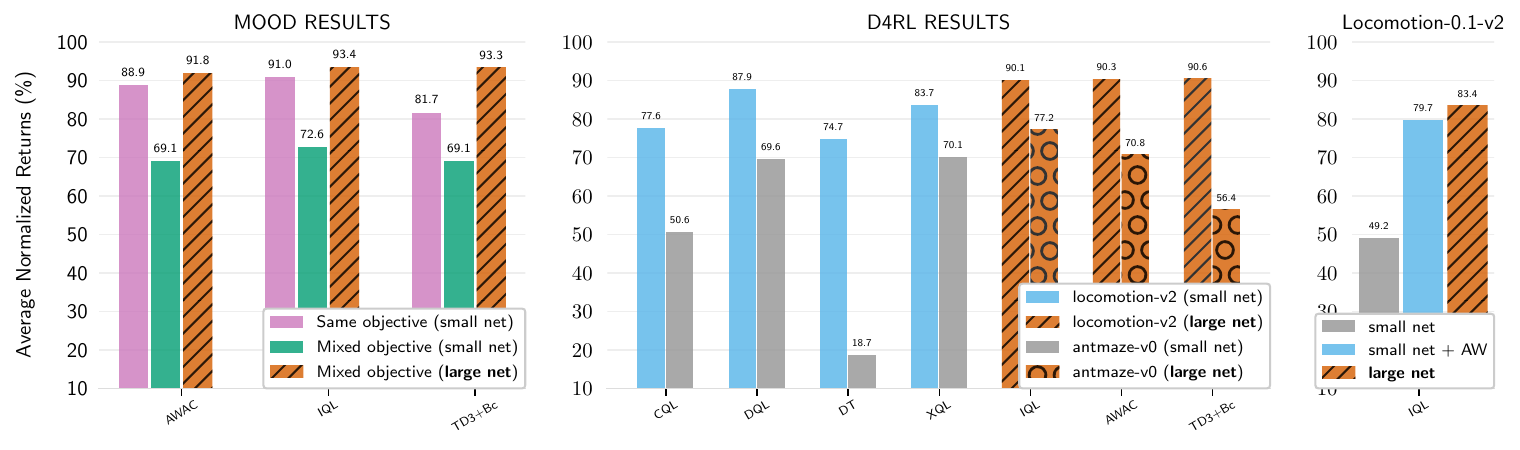}
  \caption{Average performance on the same- and mixed-objective datasets from \benchacro (\emph{left}), and the locomotion
  and antmaze datasets from D4RL (\emph{right}). The large networks are simple MLPs for MOOD and modern architectures \citep{deeper-deep-RL} for D4RL, and all involve an ensemble of 5 critics (Sec.~\ref{sec:4conjectures}). ``AW'' in the last plot denotes the sampling strategy of \citet{HongACL23harnessing} for unbalanced data.}
  \label{sec1:fig:summ_bar_perf}
\end{figure*}

However, current offline RL methods still fail in simple
    settings. \citet{Hong2023beyond,trajectory_reweighting_rel} showed that if the data contains many low-return and few high-return trajectories, policy constrained methods are unnecessarily conservative and fail to learn good behavior. \citet{heteroskedastic_rel} report a similar effect on heteroskedastic datasets where the variability of behaviors differs across different regions of the state space.
    
Realistic scenarios often involve data coming from many heterogeneous sources, such as agents trained for different tasks or demonstrations of diverse behaviors \citep{aw-opt, mocapact}. Despite the richness of these data, we show, through a novel testbed (MOOD), that existing offline RL methods can still fail: simply concatenating datasets collected from different tasks significantly and consistently \textit{hurts} performance. Counter-intuitively, this happens even for tasks where \emph{training succeeds on any of the individual subsets}, as exemplified in Fig.~\ref{sec3:fig:hum_stand_cross} with the Humanoid environment: any of the datasets is enough to learn to stand, but their union is not and leads to near-zero performance.

In light of this observation, our contributions are as follows. \textbf{1)}
  We introduce \benchacro, a new testbed for offline RL based on the DeepMind Control suite
  \citep{dmc} which involves datasets with mixed data from different
  behaviors. We use it to illustrate the negative impact of data
  diversity on offline RL methods. 
\textbf{2)} We formulate several hypotheses on the limitations that lead to such a negative result, including over-conservatism of the algorithm, scale, variance, and epistemic uncertainty, while proposing principled solutions to mitigate them. 

\textbf{3)} Through a systematic empirical analysis, we test these hypotheses and solutions across three representative algorithms (TD3+BC, AWAC, IQL) and various hyperparameters, conducting over 50,000 experiments. Surprisingly, we find that scale emerges as the key factor impacting performance: a simple increase in the number of hidden layers and units in the actor and critic architecture significantly improves the performance of all candidate algorithms on the diverse data in MOOD.
\textbf{4)} We show similar positive results in the canonical D4RL
benchmark, where AWAC and IQL, with increased network sizes,
\emph{surpass state-of-the-art performance} on the locomotion-v2 and
antmaze-v0 datasets (Fig.~\ref{sec1:fig:summ_bar_perf}), and \emph{match the performance of sophisticated sampling strategies} on the unbalanced variants of the locomotion-v2 datasets \citep{HongACL23harnessing}.


\section{Preliminaries} \label{sec:background}
Reinforcement learning problems are typically modeled by a Markov Decision
Process \citep[MDP,][]{mdp}, i.e., a tuple $(S, A, P, p_0, r, \gamma)$ with a state space $S$, an action space $A$, transition
dynamics $P: S \times A \rightarrow \mathrm{Prob}(S)$, initial state distribution $p_0 \in \mathrm{Prob}(S)$, reward function $r : S \times A \rightarrow \mathbb{R}$, and discount factor $\gamma \in [0,1)$. The goal of
an RL problem is to learn an optimal policy $\pi^\star : S \rightarrow \mathrm{Prob}(A)$, which maximizes the
expected sum of discounted rewards (a.k.a. the return): $\pi^\star \in \argmax_{\pi} \mathbb{E}^\pi[ \sum_{t=0}^\infty \gamma^t r(s_t,a_t)]$, where the expectation is under trajectories $\tau=(s_0, a_0, s_t, \dots)$ with $s_0 \sim p_0$, $a_t \sim \pi(s_t)$, and $s_{t+1}\sim P(s_t,a_t)$ for all $t\geq 0$. The algorithms analyzed in this paper are based
on the popular off-policy actor-critic framework for continuous control
\citep{dpg, ddpg}. To optimize a parameterized policy
$\pi_\theta$ (i.e., the \emph{actor}), these algorithms learn \emph{critic}
models to approximate its action-value function
$Q^{\pi_\theta}(s,a) := \mathbb{E}^{\pi_\theta}[\sum_{t=0}^\infty \gamma^t r(s_t,a_t) \mid s_0=s,a_0=a ]$,
the value function $V^{\pi_\theta}(s) := \mathbb{E}_{a\sim\pi_\theta(s)} [Q^{\pi_\theta}(s,a)]$, or the advantage function $A^{\pi_\theta}(s,a) := Q^{\pi_\theta}(s,a) - V^{\pi_\theta}(s)$. The algorithms we consider build on top of TD3 \citep{td3}, which models the critic with two randomly-initialized Q-functions $(Q_{\phi_1}, Q_{\phi_2})$. The parameters $\phi \in \{\phi_1,\phi_2\}$ of each Q-function are optimized independently via temporal difference (TD) on a dataset $\calB$ as
\begin{align}
    \label{eq:sec2:q_fn_obj}
    \argmin_\phi \E_{(s, a, s', r) \sim \calB}\left[(Q_\phi(s, a) - y)^2\right],
\end{align}
where $y = r + \gamma\E_{a'\sim \pi_\theta(s')}\left[\min(Q_{\bar\phi_1}(s', a'), Q_{\bar\phi_2}(s', a'))\right]$ is the TD target and $(\bar\phi_1,\bar\phi_2)$ are delayed versions of the parameters $(\phi_1,\phi_2)$ used to stabilize training.
The policy is then optimized in alternation with the Q-functions as
\begin{equation}
\label{eq:sec2:pi_pg_obj_cc}
\argmax_\theta\E_{s\sim \calB, a\sim \pi_\theta(s)}\left[Q_{\phi_1}(s,a)\right].
\end{equation}
These two optimization steps are commonly referred to as \textit{policy
evaluation} and \textit{policy improvement}. While in the canonical
online RL setting the agent iteratively alternates learning $\phi$ and
$\theta$ with collecting data in the environment, in offline RL
the buffer $\calB$ is collected apriori with some unknown
behavior policy $\pi_\calB$ and no further interaction with the environment is allowed. In this case, it is well known that applying off-policy
algorithms out of the box is prone to instabilities and several
modifications have been proposed to counteract their natural tendency to extrapolate beyond the provided data \citep{bear, bcq}.

\subsection{Offline RL algorithms with policy constraints} \label{sec: offline baseline}
We describe the offline RL methods employed in our analyses: TD3 + Behavior Cloning \citep[TD3+BC,][]{td3_bc}, Advantage Weighted Actor Critic
\citep[AWAC,][]{awac}, and Implicit Q-learning \citep[IQL,][]{IQL}. We focus on these specific approaches due to their simplicity and popularity: they all build on top of TD3, a state-of-the-art algorithm for off-policy RL, while adding incremental levels of conservatism in its actor and critic components. 

\textbf{TD3+BC.} TD3+BC minimally deviates from TD3, by adding a behavioral cloning term to the policy improvement objective of Equation~\ref{eq:sec2:pi_pg_obj_cc}:
\begin{equation*}
\argmax_\theta\E_{a, s\sim \calB, a'\sim \pi_\theta(s)}\left[Q_{\phi_1} (s,a') + \alpha \overline{Q} \log \pi_\theta(a|s) \right],
\end{equation*}
where $\overline{Q}$ is the absolute Q-value averaged over each minibatch
and $\alpha$ is a scaling hyper-parameter.
 Note that maximizing $\log \pi_\theta(a|s)$ with actions from $\calB$ corresponds to minimizing the forward KL divergence $D_{\mathrm{KL}}(\pi_\calB(\cdot|s)||\pi_\theta(\cdot|s))$.

\textbf{AWAC.} Similarly to TD3+BC, AWAC keeps the same critic update \eqref{eq:sec2:q_fn_obj} as TD3 while modifying the actor update \eqref{eq:sec2:pi_pg_obj_cc} as
\begin{gather}
\label{eq:awac_obj}
    \argmax_\theta\E_{a, s\sim \calB}\left[\frac{\exp(A_{\phi_1}(s,a)/\beta)}{Z}\log \pi_\theta(a|s)\right],
\end{gather}
where $A_{\phi_1}(s,a)=Q_{\phi_1}(s, a) - \E_{a'\sim\pi_\theta(s)}[Q_{\phi_1}(s, a')]$ and $Z = \E_{s, a\sim \calB }\left[\exp(A_{\phi_1}(s,a)/\beta)\right]$, while $\beta$ is a temperature hyper-parameter. In tabular settings, this is equivalent to to minimizing $D_{\mathrm{KL}}(\pi^\star_\calB(\cdot|s)||\pi_\theta(\cdot|s))$, where $\pi^\star_\calB$ is the policy maximizing the advantage $A_{\phi_1}(s,a)$ subject to an inverse KL constraint forcing $D_{\mathrm{KL}}(\pi^\star_\calB(\cdot|s)||\pi_\calB(\cdot|s)) \leq \eps$ \citep{awr_sem0, awr}.

\textbf{IQL.} IQL keeps the policy improvement \eqref{eq:awac_obj} of AWAC,
but modifies its critic to learn a parametric model of the value function $V_\psi$ using expectile regression:
\begin{equation} \label{eq:sec3:fit_v_expectiles}
        \argmin_\psi
        \E_{s, a\sim \calB}[L_2^\tau(Q(s,a) - V_\psi(s))], 
\end{equation}
where $Q(s,a) := \min(Q_{\bar\phi_1}(s, a), Q_{\bar\phi_2}(s, a))$, $L_2^\tau(u) = |\tau-1_{\text{if } (u < 0)}|u^2$, and $\tau \in (0,1)$ is a hyper-parameter. It then learns the action-value functions $(Q_{\phi_1}, Q_{\phi_2})$ by modifying the TD targets in \eqref{eq:sec2:q_fn_obj} as $y = r + \gamma V_\psi(s')$. The main advantage over the critic update of TD3 is that IQL never queries the learned Q-functions on actions outside the dataset.


\section{Offline RL with Diverse Data}

\label{sec:3evaluation_mood}

Prior offline RL methods have been extensively tested and validated
using well-known benchmarks such as D4RL~\citep{d4rl} and
RL-unplugged~\citep{rl_unplugged}, but the datasets within these benchmarks
exhibit a significant bias towards the specific task for which each
method is evaluated. Recent works \citep{Hong2023beyond,trajectory_reweighting_rel} showed that these methods tend to fail when the dataset is unbalanced (e.g., when most trajectories have low return). Here we provide a complementary analysis to highlight the challenges of incorporating diverse data sources. To this end, we introduce \benchname (\benchacro), a new testbed for offline RL to focus on this relevant problem dimension.

\subsection{The \benchacro testbed}\label{subsec:3.1:benchmark}

We build \benchacro on top of the DeepMind Control suite \citep{dmc}, spanning four environments (15 total tasks) of
increasing complexity (cheetah, walker, quadruped, and humanoid, see Fig.~\ref{sec3:fig:mood_envs} in App.~\ref{appA:bench_det}) with several mixed- or cross-task setups for each.
For each environment, we first collect data by training behavior policies for different
objectives (see Tab.~\ref{appA:tab:mood_data_coll} in App.~\ref{appA:bench_det}), including both traditional reward maximization on DMC tasks (e.g., walk, run, or stand) and the exploration-focused intrinsic motivation from Random Network
Distillation \citep[RND,][]{RND}. We train agents for several million
steps based on the difficulty of the environment, and gather
data by randomly sub-sampling $10\%$ of the resulting replay buffers. We
then merge the data coming from some subset of tasks, and relabel them for a possibly
different target task, thus building several datasets for benchmarking
offline RL methods. Each \benchacro dataset is denoted as ``\emph{domain
source-tasks $\rightarrow$ target-task}'', where ``\emph{domain}'' is the
considered environment (e.g., Walker), ``\emph{source-tasks}'' lists the
objectives whose data was merged, and ``\emph{target-task}'' is the task
used to relabel the rewards (i.e., the task to be solved on this dataset). Depending on the source and target tasks, we obtain different classes of datasets (see App.~\ref{appA:bench_det} for the details):
\setlength{\leftmargini}{15pt}
\begin{itemize}
    \setlength\itemsep{0.01em}
    \item \textit{Same-objective datasets} involve a single source task equal to the target task, akin to traditional offline benchmarks (e.g., \textit{Humanoid stand $\rightarrow$ stand} in Fig.~\ref{sec3:fig:hum_stand_cross}).
    \item \textit{Cross-objective datasets} involve a single source task which is different from the target task (e.g., \textit{Humanoid walk $\rightarrow$ stand} in Fig.~\ref{sec3:fig:hum_stand_cross}).
    \item In \textit{mixed-objective datasets}, the source tasks include all the tasks available in the chosen environment plus optionally RND (e.g., \textit{Walker mixed[+RND] $\rightarrow$ walk}).
\end{itemize}

\begin{table}[t]
    \footnotesize
    \centering
    \adjustbox{max width=0.99\linewidth}{
        \begin{tabular}{@{}lcccc@{}}
            \toprule
            Env/Algorithm & IQL & AWAC & TD3 & TD3+BC \\
            \midrule
            \multicolumn{1}{c}{} & \multicolumn{4}{c}{Same objective datasets}\\
            \cmidrule(l){2-5} 
            cheetah & $95.1\pm 1.1$ & $97.2\pm 0.8$ & $58.9\pm 6.6$ & $95.7\pm 1.3$ \\
            humanoid & $74.8\pm 3.1$ & $70.6\pm 3.3$ & $3.5\pm 0.9$ & $47.1\pm 7.7$ \\
            quadruped & $94.5\pm 0.6$ & $91.9\pm 1.7$ & $45.5\pm 3.7$ & $87.3\pm 3.0$ \\
            walker & $95.8\pm 0.9$ & $95.8\pm 0.9$ & $76.5\pm 3.5$ & $96.5\pm 0.6$ \\
            \midrule
            \multicolumn{1}{c}{\textbf{Total}} & $360.3$ & $355.5$ & $184.4$ & $326.7$ \\
            \midrule
            \multicolumn{1}{c}{} & \multicolumn{4}{c}{Mixed objective datasets}\\
                \cmidrule(l){2-5} 
                cheetah & $79.1\pm 5.1$ & $91.1\pm 2.5$ & $81.4\pm 6.3$ & $88.1\pm 4.0$ \\
                humanoid & $26.1\pm 3.2$ & $18.2\pm 4.0$ & $3.9\pm 0.7$ & $11.7\pm 3.2$ \\
                quadruped & $88.7\pm 1.1$ & $74.8\pm 1.9$ & $61.6\pm 3.7$ & $80.8\pm 2.6$ \\
                walker & $90.9\pm 2.3$ & $92.2\pm 2.0$ & $92.9\pm 2.6$ & $95.5\pm 1.6$ \\
                \midrule
                \multicolumn{1}{c}{\textbf{Total}} & $284.8$ & $276.4$ & $239.8$ & $276.2$ \\
                \midrule
                \multicolumn{1}{c}{\textbf{Average drop}}  & $-26.5$\% & $-28.65$\% & $23.09$\% & $-18.26$\% \\ 
            \bottomrule
        \end{tabular}
    }
    \caption{Average performance (plus/minus standard error) over all tasks per environment on the \benchacro datasets. All algorithms use the shallow architecture (2 hidden layers of 256 units) commonly employed in the literature. ``Average drop'' reports the performance decrease from same-objective to mixed-objective data.
    }\label{tab:results-shallow}
\end{table}

\subsection{The paradoxes of incorporating diverse data}

We use \benchacro to test our candidate offline RL methods (TD3+BC, AWAC,
and IQL) and highlight how they struggle with increasing data diversity.
We use a shallow network architecture (2 hidden layers of 256 units with ReLUs in between) for both the actor and the critic of all algorithms, as it is common in existing implementations. For each experiment, (i.e., pair of algorithm and
dataset), we perform a grid search over the hyperparameters specific
to each offline algorithm using 5 random seeds,
and select the configurations that lead to the highest cumulative return after $1.5 \times 10^6$ optimization steps ($5\times10^6$ for humanoid). We report the performance of each algorithm as the
average cumulative return normalized by the highest return of a
trajectory present in the dataset. We show in Table~\ref{tab:results-shallow} the results for the
same- and mixed-objective datasets\footnote{All mixed-objective datasets used in the main paper contain RND data except for humanoid, where the complexity of the domain makes RND generate useless samples.} averaged over all tasks in each environment. See App.~\ref{app:all-results} for all the details and results.

\textbf{Offline RL struggles with increased data diversity}. When
examining the ability of offline RL methods to leverage auxiliary data
from different tasks on the mixed-objective datasets, we consistently
observe a counter-intuitive phenomenon: \emph{adding} data from
various sources significantly reduces the performance of all the
considered offline RL algorithms. Note that no subsampling
occurs when merging the task-specific datasets: the mixed-objective dataset
is a superset of the same-objective dataset on which the algorithms work seamlessly. This phenomenon seems more pronounced in harder tasks, with a performance drop higher than $50\%$ for all offline RL algorithms in Humanoid.

\textbf{TD3 benefits from increased data diversity}. On the contrary, the performance of plain TD3 improves significantly with more diverse data. This is not surprising, as it is known that ``exploratory data'', such as the one generated by RND, allows non-conservative algorithms like TD3 to counteract their extrapolation tendency and achieve decent performance in many of the considered tasks \citep{yarats2022don}. It is thus natural to wonder why offline RL algorithms incur the opposite behavior. We formulate several hypotheses on why this happens in the next section.


\section{On the Failure of Existing Algorithms} \label{sec:4conjectures}

We list several hypotheses on why existing offline RL
methods struggle with the mixed-objective data in MOOD. For each of
them, we propose simple remedies that can be seamlessly integrated
without altering the nature of the method itself. We empirically test each of these hypotheses in Sec.~\ref{sec:5:evaluation}.

\subsection*{Hypothesis 1: over-conservatism}

The most natural cause one may think of is over-conservatism: all the considered methods force the learned policy to stay close to the data distribution. This is clearly beneficial when the data contains mostly high-return trajectories for the desired task (e.g., in D4RL or MOOD same-objective datasets), but it can have a detrimental effect when this is not the case (e.g., in MOOD mixed-objective datasets). In fact, as the data contains behaviors far from the desired one (e.g., a humanoid running or walking when the task is to stand), the learned policy may be forced to put probability mass over poor actions, hence drifting from optimality. This phenomenon was observed in recent works on unbalanced datasets containing mostly low-return trajectories \citep{Hong2023beyond,trajectory_reweighting_rel}. Other works also observed that, while constraining or regularizing the policy stabilizes training, it may degrade the evaluation performance~\citep{bear, heteroskedastic_rel,DBLP:conf/aaai/YuYSNE23}.

\begin{figure}
\begin{center}
\begin{minipage}{0.45\linewidth}
    \centering
    \includegraphics[width=1\linewidth]{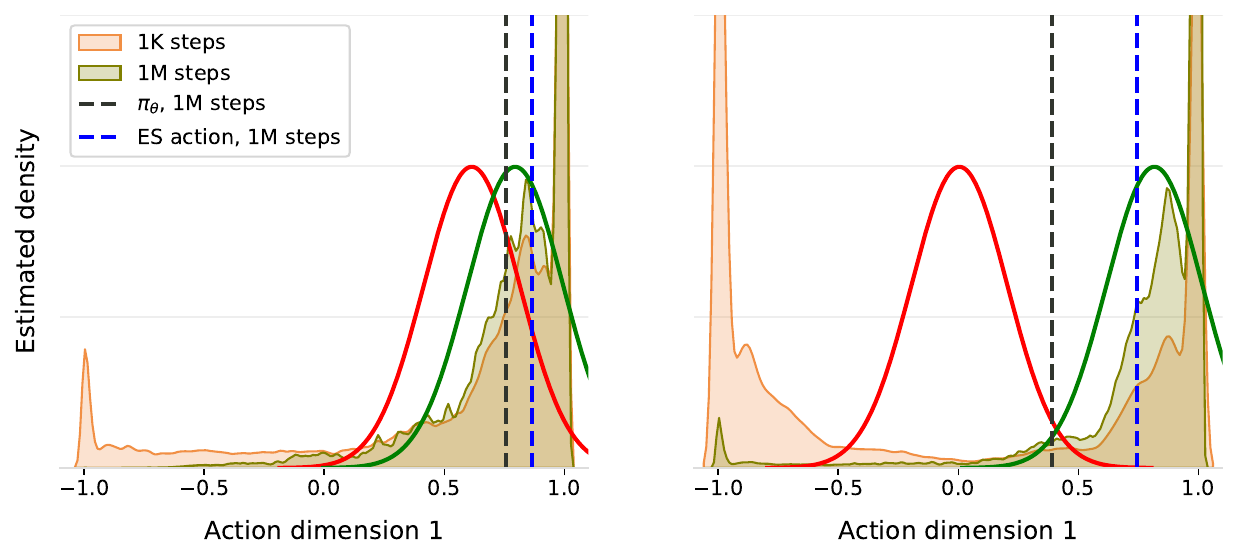}
    \caption{Optimal advantage-weighted distribution $\pi_\calB^\star$ (shaded areas) and its Gaussian projection (solid curves) after 1K and 1M optimization steps of AWAC on cheetah run with the same-objective (\emph{left}) and the mixed-objective (\emph{right}) datasets. Dashed lines indicate the actions chosen during evaluation using either the mean of the learned policy (black) or ES (blue) after 1M steps. Distributions are plotted for a randomly-chosen state and action dimension.}
    \label{sec3:fig:dist_evo_st_mt}
\end{minipage}
\hspace{0.5cm}
\begin{minipage}{0.45\linewidth}
    \centering
        \includegraphics[width=1\linewidth]{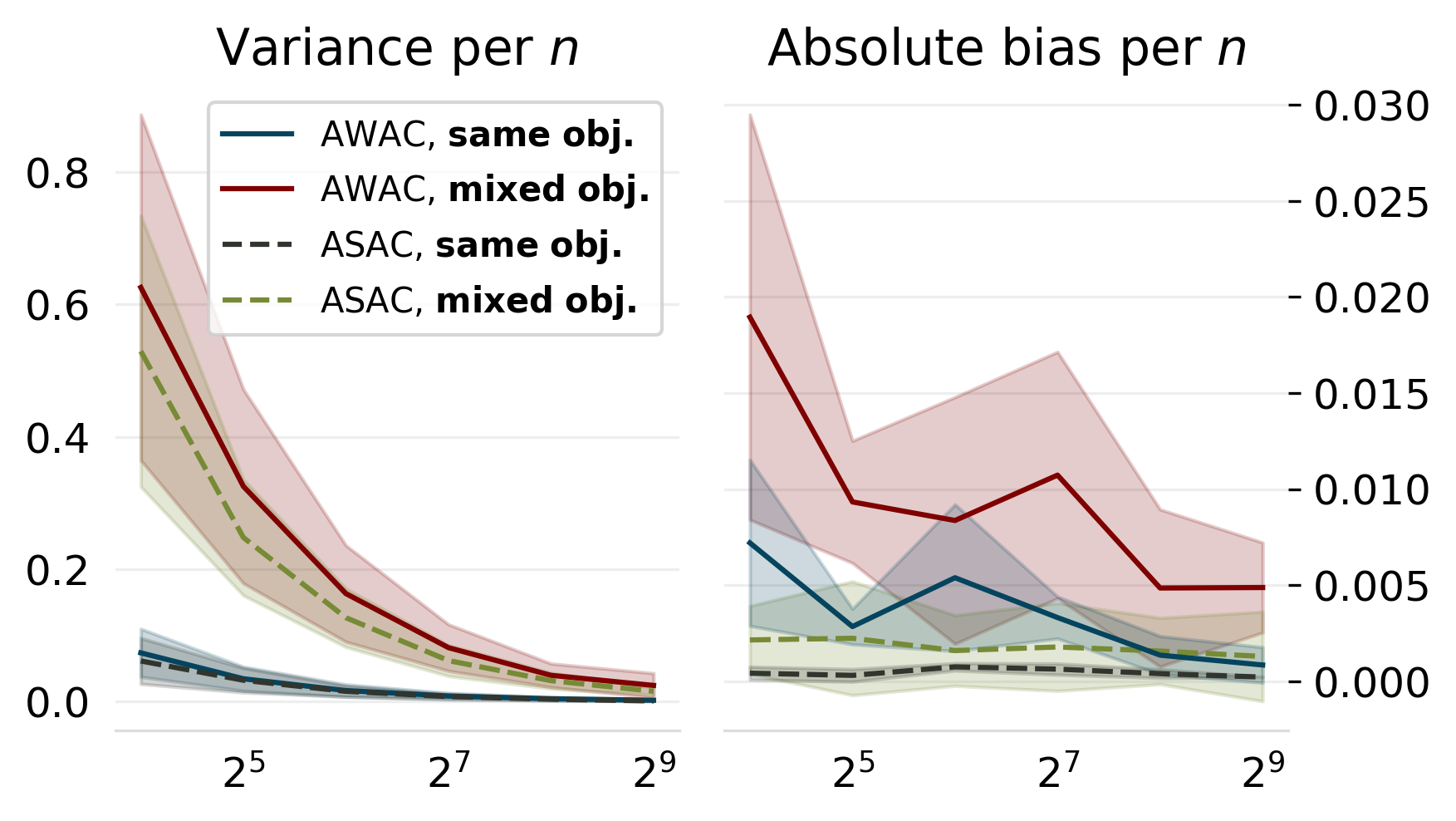}
      \caption{Empirical bias and variance of the AWAC's and \implacro's objective estimators for different batch sizes $n$ on cheetah run. Each point is averaged over 1000 randomly sampled minibatches of size $n$ at equally-spaced checkpoints saved
      during training.}
      \label{fig:awac-bias-var}
\end{minipage}
\end{center}

\end{figure}


  This conservatism may be amplified in the considered algorithms, which all fit \emph{Gaussian} policies regularized by a \emph{forward KL} term to the data distribution. 
  Given that such a distribution is often multi-modal (e.g., in the mixed-objective datasets) and given the mean-seeking tendency of the forward KL divergence, the learned policy's mean is unlikely to reflect the apex of the underlying target distribution of behaviors. We illustrate this point for AWAC in Fig.~\ref{sec3:fig:dist_evo_st_mt}, where we plot the target advantage-weighted distribution $\pi^\star_\calB(a|s) \propto \exp(A_\phi(s, a)/  \beta)
  \pi_\calB(a|s)$ (estimated through an auto-regressive density model) together with the \emph{Gaussian} policy $\pi$ that minimizes
  $D_\text{KL}(\pi^\star_\calB(\cdot|s)||\pi(\cdot|s ))$ (i.e., the policy we hope the algorithm to learn). The plots clearly show that, on the mixed-objective dataset, the learned policy's mean is very far from the target distribution's mean\footnote{While the fact that Gaussian policies poorly fit the target distribution was initially another hypothesis behind over-conservatism, we discarded it as we found such policies to have a useful regularization effect, and the usage of more expressive distributions or reverse (mode-seeking) KL led to performance collapse.}.

  \textbf{Evaluation sampling (ES).} If over-conservatism is really an issue, we propose to address it entirely at \emph{test time} by sampling $M$ actions from the learned policy and selecting the one with the highest $Q$-value, thus performing a non-parametric step of unconstrained policy improvement to skew the action distribution towards higher performance. We call this approach \emph{evaluation sampling} (ES). See Alg.~\ref{sec4:alg:on_depl} and Fig.~\ref{sec3:fig:dist_evo_st_mt}. In contrast to similar approaches \citep{CRR,EmaQ}, ES does not alter training at all, thus preserving the desired policy support constraints. We expect ES to yield positive signal mostly when the whole policy distribution is within the data support (i.e., when the policy is really over-conservative as hypothesized), as otherwise extrapolating beyond it may hinder performance.

\begin{center}
  \begin{algorithm}[tb]
    \centering
    \small
    \caption{Online deployment with evaluation sampling}
    \label{sec4:alg:on_depl}
 \begin{algorithmic}
    \STATE {\bfseries Input:} actor $\pi_\theta$, critic $Q_\phi$, number of action samples $M$
    \STATE Get initial state: $s \sim p_0$
    \WHILE{not done}
    \STATE Sample actions: $a_1, \dots a_M \sim \pi_\theta(\cdot|s)$
    \STATE $a^\star \gets \argmax_{ a\in \{a_1, \dots, a_M\} } Q_\phi(s, a)$
    \STATE Play $a^\star$, get state $s\sim P(s,a^\star)$ and reward $r(s,a^\star)$
    \ENDWHILE
 \end{algorithmic}
 \end{algorithm}
\end{center}

\subsection*{Hypothesis 2: network scale}

As the mixed-objective datasets are a strict superset of their
same-objective counterparts, both the actor and critic networks are
required to model wider regions of the state-action space. This may lead
the networks to spend capacity in modeling unnecessary quantities, as
some of these regions may actually be useless for the task at hand (e.g.,
it is not strictly necessary to model the action values of states
corresponding to the humanoid running when learning how to stand). Thus,
with networks of limited capacity, one may expect a loss of accuracy in
regions that are actually important for learning the given task. We thus
hypothesize that network scale may be one of the factors leading to the performance drop on mixed-objective data. In our experiments, we shall test this hypothesis by employing wider and deeper variants of the network architectures commonly employed in the literature, without changing any other parameter (e.g., activation functions or normalizations). 

\paragraph{Large modern architectures.} Some recent works showed that merely adopting very deep architectures can be prone to instabilities in RL \citep{andrychowicz2020matters,deeper-deep-RL,ota2021training}. To make sure we do not run into this issue, we shall also test the \emph{modern architecture} proposed by \citet{deeper-deep-RL}. Such an architecture was shown to enable stable training thanks to a combination of the fully-connected residual blocks commonly used in
transformers~\citep{transformers}
with spectral normalization~\citep{spectralNorm}. See Fig.~\ref{fig:modern} in App.~\ref{appB:impl_details}. 
This architecture was also shown to help counteract the plasticity loss of the networks \citep{contrib_pessimism}, a phenomenon that frequently occurs when training for non-stationary objectives \citep{warmstart,plasticity_loss_cl} as in RL \citep{plasticity_rl_0, plasticity_rl_1,
plasticity_rl_2,bbf_large_critic}. We conjecture that such a non-stationarity may be amplified with mixed-objective data (cf. the drift of the target policy $\pi_\calB^\star$ in Fig.~\ref{sec3:fig:dist_evo_st_mt}), and the modern architecture may also help in this regard.

\subsection*{Hypothesis 3: epistemic uncertainty}

It is well-known that off-policy actor-critic algorithms like DDPG and
TD3 are subject to the same Q-value overestimation bias as in Q-learning
with discrete actions \citep{td3}. In online RL, some level of
overestimation may be acceptable (e.g., to encourage exploration), and
the learner could always correct wrong estimates by gathering further
data from the environment. But in offline RL it is important to guarantee
pessimistic Q-value estimates for state-action pairs not sufficiently
covered by the dataset \citep{jin2021pessimism}. Special care needs to be
taken for algorithms, like TD3+BC and AWAC, that query the learned value
functions on actions outside the given dataset: these algorithms may
result in erroneous overestimations without the possibility to ever
correct them, hence yielding poor evaluation performance. This phenomenon
may be further exacerbated in mixed-objective data, where errors can
propagate due to over-generalization across different regions of the
state-action space. We thus hypothesize that the solution proposed in
TD3, consisting of training an ensemble of two independent Q-functions
with TD targets involving the minimum between them
(Eq.~\ref{eq:sec2:q_fn_obj}), may not be sufficient to counteract this
issue. Some works \citep{lan2019maxmin,chen2021randomized} indeed showed
that using a larger ensemble of critics can reduce both the Q-value
estimation bias and variance. We shall thus test this solution in our
experiments. Formally, we train an ensemble of $n$ randomly-initialized
Q-functions $(Q_{\phi_1}, \dots, Q_{\phi_n})$ while independently
optimizing their parameters via TD learning as in
\eqref{eq:sec2:q_fn_obj}. Following \citet{contrib_pessimism}, we
redefine the TD targets in \eqref{eq:sec2:q_fn_obj} as $y = r + \gamma\E_{a'\sim \pi_\theta(s')}\left[\overline{Q}(s,a)\right]$, where
\begin{align*}
    \overline{Q}(s,a) := \frac{1}{n}\sum\limits_{i} Q_{\phi_i}(s,a) - \frac{\lambda}{n^2 - n}\sum\limits_{i,j} |Q_{\phi_i}(s,a)- Q_{\phi_j}(s,a)|,
\end{align*}
with a hyperparameter $\lambda \geq 0$. This gives us more flexibility in the aggregation of these Q-functions: for $n=2$ and $\lambda=0.5$, we recover the same update rule as TD3, while for other choices of $\lambda$ we can control the level of pessimism.


\subsection*{Hypothesis 4: bias and variance of advantage weighting}

This hypothesis focuses on advantage-weighted algorithms like AWAC and IQL. We recall that the policy improvement objective of AWAC (Eq.~\ref{eq:awac_obj}) is implemented by employing a weighted importance sampling (WIS) estimator:
\begin{equation}\small \label{sec4:eq:awac_softmax}
    J^\text{AW}(\theta)=\E_{(a_{1:n},s_{1:n})\sim \calB}\left[\sum^n_{i=1}w(s_i, a_i)\log \pi_\theta(a_i|s_i)\right],
\end{equation}
where the weights $w(s_i, a_i) = \frac{\exp(A_\phi(s_i, a_i)/\beta)}{\sum^n_{j=1}\exp(A_\phi(s_j, a_j)/\beta)}$ are normalized over a minibatch of size $n$. It is known that this estimator is both biased and introduces higher variance than directly sampling from the desired target distribution \citep{weighted_is_bias_var}. On mixed-objective data, as the distributions involved become more complex, it is natural to expect bias and variance to increase (cf. Fig.~\ref{fig:awac-bias-var}). We thus hypothesize this fact to be one of the reasons behind the performance drop of the considered advantage-weighted algorithms.

If this is really the case, we propose a very simple workaround: we can directly and tractably sample from the desired target distribution by avoiding altogether the need for weights in the objective. Formally, we define a modified sampling data distribution for policy improvement:
\begin{equation}\label{sec4:eqn:prioritized_B_dist}
\calB^\star(s,a)\deq \frac{1}{Z} \calB(s,a) \exp(A_\phi(s,a)/\beta)
\end{equation}
where $\calB(s, a)$ denotes the distribution obtained by i.i.d.\ sampling from the buffer, while $Z = \sum_{s, a\in \calB}\exp(A_\phi(s, a)/\beta)$.
To sample from $\mathcal{B}^\star$ efficiently without explicitly computing $Z$, we design a \textit{logsumexp-tree} inspired by the sum-tree data structure used for prioritized sampling \citep{per}, where we store the scaled advantages $A_\phi(s,a)/\beta$ and work entirely in log-space.
Then, we simply train $\pi$ to maximize the likelihood of the data sampled from $\calB^\star$: 
\begin{equation}\label{eq:policy-obj}
\hat J^{\text{AS}}(\theta) = \E_{(s,a)\sim \calB^*} [\log \pi_\theta(a|s)].
\end{equation}
It is easy to see that this estimator, which aims at directly projecting $\pi_\theta$ onto $\pi^\star_\calB$, is \emph{unbiased} for the AWAC objective, i.e., its expectation is equal to the objective in \eqref{eq:awac_obj}.


We call the resulting approach \emph{advantage sampled actor critic}
(ASAC, see Alg.~\ref{sec4:alg:off_learn}). 
\footnote{
Alg.~\ref{sec4:alg:off_learn} only updates $\calB^\star(s, a)$
on each
minibatch rather than the whole buffer, so some stored values of $A_\phi$ may grow
``stale'': this re-introduces some bias for \eqref{eq:policy-obj}. 
Fig.~\ref{fig:awac-bias-var} shows this bias is limited.
}
Note that it bears some similarities with methods that alter the data distribution via weighting techniques~\citep{Hong2023beyond, trajectory_reweighting_rel, oper_rel}. The main difference is that these methods compute weights for offline trajectories only once at the start of training based on the returns/advantages of the behavior policy, while ASAC's sampling distribution adaptively evolves over time.

\begin{algorithm}[tb]
    \small
    \caption{Advantage Sampled Actor Critic (ASAC)}
    \label{sec4:alg:off_learn}
 \begin{algorithmic}
    \STATE {\bfseries Input:} offline data $\calB$
    \WHILE{not done}
        \STATE \emph{// Actor update}
    \STATE Sample batch $b_{\mathrm{ac}} = \{(s, a)\}$ from $\calB^\star$ (see Eq.~\ref{sec4:eqn:prioritized_B_dist})
    \STATE Take gradient step on $\theta$ to maximize \eqref{eq:policy-obj} on $b_{\mathrm{ac}}$
    \STATE \emph{// Critic update (TD3)}
    \STATE Sample batch $b_{\mathrm{cr}} = \{(s, a, s', r)\}$ from $\calB$
    \STATE Take gradient step on $\phi$ to minimize \eqref{eq:sec2:q_fn_obj} on $b_{\mathrm{cr}}$
    \STATE \emph{// Update sum-tree for $\calB^\star$}
    \STATE $\calB^\star(s, a) \gets \exp(A_\phi(s, a)/\beta)$ for all $(s,a) \in b_{\mathrm{cr}} \cup b_{\mathrm{ac}}$
    \ENDWHILE
 \end{algorithmic}
 \end{algorithm}


 \begin{table*}[t]
    \footnotesize
    \centering
    \adjustbox{max width=0.99\linewidth}{
\begin{tabular}{lcccccccccc}
    \toprule
    Env/Algorithm & \multicolumn{2}{c}{IQL} & \multicolumn{2}{c}{ASAC} & \multicolumn{2}{c}{AWAC} & \multicolumn{2}{c}{TD3} & \multicolumn{2}{c}{TD3+BC} \\
    \midrule
    Architecture (large) & Modern & Simple & Modern & Simple & Modern & Simple & Modern & Simple & Modern & Simple  \\
    \cmidrule(l){2-11} 
    \textbf{Total} (2 critics, no ES) &  $369.7$ & $373.0$ & $347.4$ & $331.3$ & $359.5$ & $355.9$ & $228.9$ & $202.5$ & $358.9$ & $370.2$ \\
    \midrule
    \textbf{Total} (5 critics, no ES) & $370.0$ & $373.6$ & $364.9$ & $355.9$ & $367.5$ & $367.3$ & $228.9$ & $207.6$ & $365.8$ & $373.1$ \\
    \bottomrule
    \midrule
        \multicolumn{1}{c}{} & \multicolumn{10}{c}{5 critics / large simple architecture}\\
        \cmidrule(l){2-11} 
        ES & \xmark & \cmark & \xmark & \cmark & \xmark & \cmark & \xmark & \cmark & \xmark & \cmark \\
        \midrule
        cheetah & $92.5\pm 1.8$ & $89.4\pm 3.4$ & $94.9\pm 1.1$ & $95.4\pm 1.4$ & $95.7\pm 1.0$ & $95.1\pm 1.5$ & $81.0\pm 6.8$ & $85.1\pm 5.4$ & $94.5\pm 1.7$ & $94.6\pm 1.9$ \\
        humanoid & $92.1\pm 1.6$ & $80.9\pm 2.0$ & $91.7\pm 1.0$ & $66.2\pm 0.5$ & $88.3\pm 1.2$ & $78.6\pm 3.4$ & $4.7\pm 1.1$ & $2.9\pm 0.9$ & $86.2\pm 1.3$ & $77.3\pm 3.6$ \\
        quadruped & $94.0\pm 0.7$ & $97.1\pm 0.3$ & $74.7\pm 1.3$ & $89.9\pm 1.1$ & $86.7\pm 2.0$ & $95.6\pm 0.7$ & $26.9\pm 2.7$ & $31.7\pm 3.6$ & $94.9\pm 0.6$ & $98.1\pm 0.5$ \\
        walker & $95.0\pm 1.3$ & $95.7\pm 1.3$ & $94.6\pm 1.5$ & $97.4\pm 0.8$ & $96.6\pm 1.1$ & $98.2\pm 0.6$ & $95.1\pm 1.9$ & $95.3\pm 1.8$ & $97.5\pm 0.3$ & $99.0\pm 0.4$ \\
        \midrule
        \multicolumn{1}{c}{\textbf{Total}} & $373.6$ & $363.2$ & $355.9$ & $348.9$ & $367.3$ & $367.4$ & $207.6$ & $215.0$ & $373.1$ & $368.9$ \\
        \midrule
        \multicolumn{1}{c}{\textbf{Total} (max over ES)} & \multicolumn{2}{c}{$378.6$} & \multicolumn{2}{c}{$372.7$} & \multicolumn{2}{c}{$378.3$} & \multicolumn{2}{c}{$218.3$} & \multicolumn{2}{c}{$378.1$} \\
        \bottomrule
        \midrule
        \multicolumn{1}{c}{} & \multicolumn{10}{c}{5 critics / large modern architecture}\\
        \cmidrule(l){2-11} 
        ES & \xmark & \cmark & \xmark & \cmark & \xmark & \cmark & \xmark & \cmark & \xmark & \cmark \\
        \midrule
        cheetah & $93.2\pm 1.6$ & $89.0\pm 3.7$ & $94.7\pm 1.3$ & $94.2\pm 1.2$ & $95.0\pm 1.3$ & $93.8\pm 1.9$ & $76.3\pm 5.5$ & $81.6\pm 6.6$ & $91.9\pm 1.9$ & $91.2\pm 3.2$ \\
        humanoid & $89.9\pm 1.8$ & $78.2\pm 2.8$ & $90.4\pm 0.8$ & $79.1\pm 2.2$ & $91.4\pm 0.6$ & $78.5\pm 3.7$ & $3.3\pm 0.4$ & $3.0\pm 0.5$ & $83.5\pm 1.4$ & $76.4\pm 3.9$ \\
        quadruped & $92.1\pm 0.8$ & $96.7\pm 0.3$ & $83.2\pm 1.5$ & $92.2\pm 0.8$ & $85.4\pm 2.0$ & $92.2\pm 1.5$ & $56.1\pm 2.0$ & $57.9\pm 3.0$ & $92.1\pm 0.9$ & $93.8\pm 0.9$ \\
        walker & $94.6\pm 1.4$ & $95.3\pm 1.3$ & $96.5\pm 0.5$ & $96.7\pm 0.7$ & $95.8\pm 1.1$ & $95.5\pm 1.5$ & $93.1\pm 1.9$ & $94.8\pm 2.1$ & $98.2\pm 0.4$ & $98.1\pm 0.5$ \\
    \midrule
    \multicolumn{1}{c}{\textbf{Total}} & $370.0$ & $359.2$ & $364.9$ & $362.2$ & $367.5$ & $360.0$ & $228.9$ & $237.4$ & $365.8$ & $359.4$ \\
    \midrule
    \multicolumn{1}{c}{\textbf{Total} (max over ES)} & \multicolumn{2}{c}{$376.3$} & \multicolumn{2}{c}{$377.6$} & \multicolumn{2}{c}{$377.2$} & \multicolumn{2}{c}{$242.0$} & \multicolumn{2}{c}{$368.1$} \\
    \bottomrule
        \end{tabular}
    }
    \caption{Average performance over all tasks per environment on the mixed-objective MOOD datasets, with totals denoting their sum over environments. The table on top compares the \emph{large} simple and modern architectures when using an ensemble of either 2 or 5 critics. The tables on the bottom compare these architectures with 5 critics when using or not ES (with $M=50$ samples).}\label{tab:mood-all-results}
\end{table*}
\begin{table*}[t]
    \footnotesize
\centering
\adjustbox{max width=0.98\linewidth}{
\begin{tabular}{lcccccccc}
    \toprule
    agent & \multicolumn{2}{c}{IQL} & \multicolumn{2}{c}{ASAC} & \multicolumn{2}{c}{AWAC} & \multicolumn{2}{c}{TD3+BC} \\
    \midrule
    ES & \xmark & \cmark & \xmark & \cmark & \xmark & \cmark & \xmark & \cmark \\
    \midrule
    Antmaze-v0 (5 critics) & 65.2\textpm 3.5 & 77.2\textpm 2.8 & 64.5\textpm 4.1 & 72.6\textpm 3.8 & 64.7\textpm 4.1 & 70.8\textpm 4.0 & 50.7\textpm 5.9 & 51.9\textpm 7.0 \\
    Antmaze-v0 (10 critics) & 63.4\textpm 3.3 & 76.0\textpm 2.8 & 65.3\textpm 4.0 & 72.6\textpm 3.6 & 69.8\textpm 4.4 & 72.6\textpm 4.0 & 49.8\textpm 7.1 & 44.2\textpm 7.6 \\
    Locomotion-v2 (5 critics) & 88.1\textpm 3.4 & 72.3\textpm 4.3 & 84.1\textpm 3.2 & 80.7\textpm 3.5 & 87.0\textpm 3.3 & 81.6\textpm 3.3 & 82.6\textpm 2.9 & 84.9\textpm 2.5 \\
    Locomotion-0.1-v2 (10 critics) & 83.1\textpm 4.2 & 50.6\textpm 4.6 & 64.1\textpm 4.0 & 48.6\textpm 4.0 & 54.6\textpm 4.5 & 51.7\textpm 4.2 & 49.6\textpm 4.5 & 45.5\textpm 3.8 \\
    \bottomrule
\end{tabular}
}
\caption{Average performance across all datasets in each category of D4RL. All algorithms use the large modern architecture.}
\label{tab:d4rl-es}
\end{table*}

\section{Empirical Results} \label{sec:5:evaluation}

We systematically evaluate three candidate algorithms (TD3+BC, AWAC, and IQL) combined
with
the algorithmic design considerations from
Sec.~\ref{sec:4conjectures}: the use of a large simple MLP network, a
large modern architecture, an ensemble of critics, evaluation sampling, and advantage sampling.

We use deeper architectures for the actor compared to the critic, as we found the former to be more important.
For the large simple architecture, we employ critics with 3 hidden layers of 256 units each and actors with 5 hidden layers of 1024 units each. For the modern architecture, we use critics with a single modern block (cf. Fig.~\ref{fig:modern}) with 256 as hidden dimension and actors with 2 modern blocks with 1024 as hidden dimension. This makes the simple and modern networks of equal capacity, disregarding the marginal increase in parameters introduced by layer normalization.

For each design choice and task, we perform a hyperparameter sweep over the learning rate, the temperature $\beta$ (for AWAC and IQL), the regularization strength $\alpha$ (for TD3+BC), and the expectile $\tau$ (for IQL). We then report the cumulative return of the best configurations averaged over 5 random seeds. Due to space constraints, we report only the minimal results needed to justify our claims, while referring the reader to App.~\ref{app:all-results} for the complete evaluation.

\subsection{\benchacro evaluation}

Table~\ref{tab:mood-all-results} reports the results on the
mixed-objective datasets in MOOD. We notably observe that all algorithms
bridge the performance gap with same-objective datasets (cf
Tab.~\ref{tab:results-shallow}) simply by using a larger architecture,
while all other conjectured solutions either marginally help or do not
help at all. This makes us conclude that, among our hypotheses, scale is
the key factor affecting the performance drop. Interestingly, the modern
architecture does not appear to provide any advantage over the simple
one. ASAC's performance is on par with AWAC, indicating that the variance
of the AWAC estimator is not a limiting factor. Increasing the number of
critics appears to yield a consistent, though marginal, performance improvement, but at the cost of added compute. 

To a lesser extent than scale, ES contributes to performance improvements, particularly in Quadruped, but it has a detrimental effect on Humanoid. This discrepancy may be attributed to the availability of RND data with good coverage for the non-humanoid tasks, which makes it easier for conservative methods to skew the whole policy distribution within the support of the data, hence enabling ES to safely improve performance. This is not the case on the Humanoid datasets comprising only ``purposeful'' trajectories, where extrapolation has actually a detrimental effect. 
Further evidence for this conjecture is given in App.~\ref{app:exorl}, where we show that ES consistently improves performance on pure RND data from ExoRL \citep{yarats2022don}.



\begin{table*}[t]
    \footnotesize
\tabcolsep=0.04cm
\centering
\adjustbox{max width=0.98\linewidth}{
    \begin{tabular}{l|cccc|cccccccc}
        \toprule
        Dataset/Algorithm & DT & CQL & XQL & DQL & \multicolumn{2}{c}{IQL} & \multicolumn{2}{c}{ASAC} & \multicolumn{2}{c}{AWAC} & \multicolumn{2}{c}{TD3+BC} \\
        Architecture & orig. & orig. & orig. & orig. & modern & simple & modern & simple & modern & simple & modern & simple \\
        \midrule
        halfcheetah-medium-expert-v2 & 86.8 & 91.6 & 94.2 & 96.8 & 93.4\textpm 0.1 & 93.5\textpm 0.2 & 94.5\textpm 0.2 & 88.5\textpm 1.2 & 94.3\textpm 0.4 & \textbf{100.3\textpm 1.0} & 92.7\textpm 0.2 & 94.1\textpm 0.8 \\
        halfcheetah-medium-replay-v2 & 36.6 & 45.5 & 45.2 & 47.8 & 49.9\textpm 0.1 & 52.0\textpm 0.2 & 54.5\textpm 0.2 & 54.9\textpm 0.3 & 54.4\textpm 0.3 & 57.5\textpm 0.1 & 57.9\textpm 0.1 & \textbf{59.1\textpm 0.6} \\
        halfcheetah-medium-v2 & 42.6 & 44.0 & 48.3 & 51.1 & 58.2\textpm 0.1 & 63.0\textpm 0.1 & 61.3\textpm 0.5 & 61.1\textpm 0.2 & 61.4\textpm 0.1 & 64.6\textpm 0.2 & 64.6\textpm 0.3 & \textbf{70.8\textpm 0.4} \\
        hopper-medium-expert-v2 & 107.6 & 105.4 & 111.2 & 111.1 & 110.5\textpm 0.2 & 110.0\textpm 0.2 & 110.9\textpm 0.2 & 109.8\textpm 0.2 & 110.3\textpm 0.1 & 110.4\textpm 0.2 & \textbf{111.8\textpm 0.3} & 109.6\textpm 0.4 \\
        hopper-medium-replay-v2 & 82.7 & 95.0 & 100.7 & 101.3 & 101.1\textpm 0.5 & 94.7\textpm 1.9 & 101.4\textpm 0.4 & 95.2\textpm 1.4 & 101.9\textpm 0.6 & \textbf{102.5\textpm 0.5} & 100.6\textpm 0.2 & 98.7\textpm 2.1 \\
        hopper-medium-v2 & 67.6 & 58.5 & 74.2 & 90.5 & \textbf{98.1\textpm 0.8} & 96.4\textpm 2.0 & 94.6\textpm 4.7 & 75.6\textpm 1.6 & 96.4\textpm 2.8 & 81.7\textpm 16.2 & 94.9\textpm 0.8 & 93.2\textpm 0.7 \\
        walker2d-medium-expert-v2 & 108.1 & 108.8 & 112.7 & 109.6 & 112.9\textpm 0.3 & 114.1\textpm 0.4 & 112.1\textpm 0.2 & 113.5\textpm 0.3 & 112.8\textpm 0.1 & \textbf{115.5\textpm 0.3} & \textbf{115.0\textpm 0.4} & \textbf{115.5\textpm 0.9} \\
        walker2d-medium-replay-v2 & 66.6 & 77.2 & 82.2 & 95.5 & \textbf{96.2\textpm 0.2} & 94.2\textpm 0.7 & 93.9\textpm 0.5 & 91.1\textpm 1.7 & 94.7\textpm 0.4 & \textbf{96.3\textpm 1.0} & 90.5\textpm 0.3 & 87.7\textpm 0.4 \\
        walker2d-medium-v2 & 74.0 & 72.5 & 84.2 & 87.0 & \textbf{90.4\textpm 0.2} & \textbf{89.8\textpm 0.4} & 86.1\textpm 0.1 & 84.6\textpm 0.2 & 87.1\textpm 0.1 & 87.0\textpm 0.3 & 87.1\textpm 0.2 & 87.2\textpm 0.3 \\
        \midrule
        \multicolumn{1}{c|}{\textbf{Locomotion-v2 total}} & 672.6 & 698.5 & 752.9 & 790.7 & $810.7$ & $807.7$ & $809.3$ & $774.3$ & $813.1$ & \textbf{815.7} & $815.0$ & $816.0$ \\
        \bottomrule
        antmaze-large-diverse-v0 & 0.0 & 14.9 & 49.0 & 56.6 & 60.4\textpm 1.2 & \textbf{64.4\textpm 1.3} & 49.9\textpm 1.5 & 49.2\textpm 0.8 & 43.0\textpm 3.4 & 32.0\textpm 1.1 & 18.8\textpm 1.5 & 8.7\textpm 1.3 \\
        antmaze-large-play-v0 & 0.0 & 15.8 & 46.5 & 46.4 & \textbf{54.2\textpm 1.1} & 26.2\textpm 2.3 & 43.8\textpm 2.4 & 12.3\textpm 1.0 & 42.2\textpm 0.8 & 17.3\textpm 0.7 & 8.3\textpm 5.4 & 5.6\textpm 3.5 \\
        antmaze-medium-diverse-v0 & 0.0 & 53.7 & 73.6 & 78.6 & 82.9\textpm 1.1 & 85.6\textpm 0.6 & 82.3\textpm 0.8 & \textbf{89.1\textpm 1.5} & 83.2\textpm 1.5 & 85.8\textpm 1.1 & 75.0\textpm 7.5 & 61.9\textpm 14.1 \\
        antmaze-medium-play-v0 & 0.0 & 61.2 & 76.0 & 76.6 & 82.9\textpm 0.7 & \textbf{86.5\textpm 0.7} & \textbf{84.9\textpm 1.4} & \textbf{86.7\textpm 1.2} & \textbf{85.6\textpm 1.7} & 84.6\textpm 1.0 & 70.5\textpm 1.7 & 24.2\textpm 10.0 \\
        antmaze-umaze-diverse-v0 & 53.0 & 84.0 & 82.0 & 66.2 & \textbf{86.6\textpm 0.7} & 83.9\textpm 0.9 & 81.1\textpm 2.9 & 72.1\textpm 1.5 & 72.8\textpm 3.3 & 48.5\textpm 10.3 & 71.7\textpm 4.9 & 69.4\textpm 3.7 \\
        antmaze-umaze-v0 & 59.2 & 74.0 & 93.8 & 93.4 & 96.1\textpm 0.7 & 96.6\textpm 0.4 & 98.7\textpm 0.1 & \textbf{99.0\textpm 0.2} & 97.8\textpm 0.5 & 98.2\textpm 0.3 & 94.1\textpm 1.2 & 96.0\textpm 1.2 \\
        \midrule
        \multicolumn{1}{c|}{\textbf{Antmaze-v0 total}} & $112.2$ & $303.6$ & $420.9$ & $417.8$ & \textbf{463.1} & $443.2$ & $440.8$ & $408.5$ & $424.8$ & $366.3$ & $338.4$ & $265.9$ \\
        \bottomrule
    \end{tabular}
}
\adjustbox{max width=0.98\linewidth}{
    \begin{tabular}{l|ccc|cccccccc}
        \noalign{\vskip 5mm} 
        \toprule
        Dataset/Algorithm & CQL & IQL & TD3+BC & \multicolumn{2}{c}{IQL} & \multicolumn{2}{c}{ASAC} & \multicolumn{2}{c}{AWAC} & \multicolumn{2}{c}{TD3+BC} \\
        Architecture & orig. & orig. & orig. & modern & simple & modern & simple & modern & simple & modern & simple \\
        \midrule
        halfcheetah-random-expert-0.1-v2 & 45.8 & \textbf{91.3} & 78.4 & 65.8\textpm 2.2 & 76.3\textpm 2.1 & 56.2\textpm 6.6 & 61.5\textpm 6.7 & 66.7\textpm 3.0 & 78.3\textpm 1.2 & 18.3\textpm 1.2 & 30.7\textpm 1.2 \\
        halfcheetah-random-medium-0.1-v2 & 45.8 & 43.1 & 47.8 & 48.9\textpm 0.3 & 52.7\textpm 0.2 & 53.2\textpm 0.4 & 53.6\textpm 4.1 & 54.6\textpm 0.1 & 58.9\textpm 0.2 & 57.5\textpm 0.3 & \textbf{60.7\textpm 0.7} \\
        hopper-random-expert-0.1-v2 & 109.7 & \textbf{111.5} & 107.7 & 109.0\textpm 0.4 & 95.4\textpm 2.2 & 73.4\textpm 5.3 & 41.8\textpm 2.7 & 32.3\textpm 4.8 & 78.4\textpm 1.5 & 80.2\textpm 4.4 & 77.0\textpm 2.0 \\
        hopper-random-medium-0.1-v2 & 66.6 & 57.1 & 56.4 & \textbf{93.9\textpm 2.3} & 70.5\textpm 3.0 & 69.0\textpm 2.1 & 66.7\textpm 2.3 & 71.7\textpm 1.2 & 76.3\textpm 3.8 & 71.1\textpm 1.9 & 60.5\textpm 1.7 \\
        walker2d-random-expert-0.1-v2 & 108.1 & 109.3 & \textbf{110.1} & 107.8\textpm 0.3 & 99.2\textpm 0.6 & 99.5\textpm 2.0 & 101.1\textpm 0.6 & 93.6\textpm 3.1 & 61.3\textpm 15.5 & 41.9\textpm 17.0 & 11.6\textpm 0.8 \\
        walker2d-random-medium-0.1-v2 & 66.3 & 65.8 & 74.2 & 75.0\textpm 0.5 & 73.3\textpm 0.6 & 65.9\textpm 4.7 & 75.0\textpm 0.9 & 47.0\textpm 6.9 & \textbf{78.4\textpm 3.1} & 57.5\textpm 4.3 & 53.1\textpm 10.4 \\
        \midrule
        \multicolumn{1}{c|}{\textbf{Locomotion-0.1-v2 total}} & 442.3 & 478.1 & 474.6 & \textbf{500.5} & 467.4 & 417.2 & 399.8 & 365.9 & 431.5 & 326.5 & 293.5 \\
        \bottomrule
    \end{tabular}
    }
\caption{Performance on the locomotion-v2 and antmaze-v0 datasets from the D4RL benchmark (\emph{top}), and on the unbalanced variants of the locomotion-v2 datasets (\emph{bottom}). For our candidate algorithms we use \emph{large} architectures, 5 critics (10 for the unbalanced data), and report the best result between using or not ES (with $M=50$ samples). The second block of the bottom table reports the performance of the algorithms tested by \citet{HongACL23harnessing} with their modified sampling distribution (values taken from their \href{https://github.com/improbable-ai/harness-offline-rl?tab=readme-ov-file}{Github repository}).}
\label{sec5:tab:d4rl}
\end{table*}

\subsection{D4RL evaluation}

Table \ref{sec5:tab:d4rl} reports the results of IQL, AWAC, TD3+BC, and
ASAC
with large architectures, 5 critics, and choosing the best performance
between using ES or not. We also report previously-published results for
state-of-the-art algorithms: decision transformer
\citep[DT,][]{decision_transf}, conservative Q-learning
\citep[CQL,][]{cql}, extreme Q-learning \citep[XQL,][]{XQL}, and
Diffusion Q-learning~\citep[DQL,][]{WangHZ23DQL}. Overall, IQL, AWAC, and
ASAC with the large modern architecture surpass the state of the art,
while the performance of TD3+BC still falls behind in the antmaze tasks.
We emphasize that, while CQL and XQL use smaller architectures, the
purpose of these experiments is not to establish which algorithm is best,
but rather to showcase that several existing simple strategies with
increased scale are
competitive with more complicated approaches.

\textbf{Scale and diversity.} The performance gap between large and small
networks is crucial for Antmaze but marginal for Locomotion (see App.~\ref{app:all-results}).
This aligns with the observations from MOOD: D4RL
locomotion collects data with a protocol similar to same-objective MOOD,
while the Antmaze datasets, being generated by multi-goal reaching policies, involve data diversity comparable to mixed-objective MOOD.

\textbf{Unbalanced data.}  \citet{HongACL23harnessing} observed that offline RL algorithms struggle on the unbalanced locomotion datasets consisting of $90\%$ trajectories generated by the uniform policy and $10\%$ of expert or medium trajectories. They conjectured over-conservatism to be the main issue, as the performance of policy-constrained methods is tightly coupled to the (poor) behavior policy. As a solution, they suggested sampling trajectories proportionally to their cumulative return, essentially discarding the random transitions. Our results (Tab.~\ref{sec5:tab:d4rl}) suggest that this is actually not needed: simply using larger networks, while training with standard uniform sampling, achieves comparable results as theirs.

\textbf{The role of ES.} As can be noted in Tab.~\ref{tab:d4rl-es}, ES
improves the performance on antmaze tasks by a large margin, while it actually hurts in all locomotion datasets. This is likely to be due to a similar phenomenon as in the mixed-objective MOOD data: the good coverage of antmaze data enables an accurate learning of the Q-functions within the policy support, as opposed to the narrower distributions of the locomotion tasks where extrapolation is dangerous.


\section{Conclusions}
\label{sec7:conclusion}

We provided empirical evidence that scale is a key factor in overcoming some difficulties of current offline RL methods, such as training from diverse data not directly related to the target task. In our study, scale appears more important than algorithmic considerations: two simple algorithms (AWAC, IQL) surpass state-of-the-art methods on the standard D4RL benchmark, and even TD3+BC comes reasonably close. In contrast, further algorithmic refinements yield only limited benefits. This questions the common trend of designing increasingly more complicated algorithms, as even simple approaches can shine with proper scaling.


\clearpage
\newpage
\bibliographystyle{assets/plainnat}
\bibliography{main}


\clearpage
\newpage
\beginappendix
\addcontentsline{toc}{section}{Appendix} 
\part{Appendix} 
\parttoc 
\newpage
\section{\benchacro details}

\label{appA:bench_det}

\begin{wrapfigure}{r}{0.55\textwidth}
\vspace{-6mm}
\begin{center}
    \includegraphics[width=0.55\textwidth]{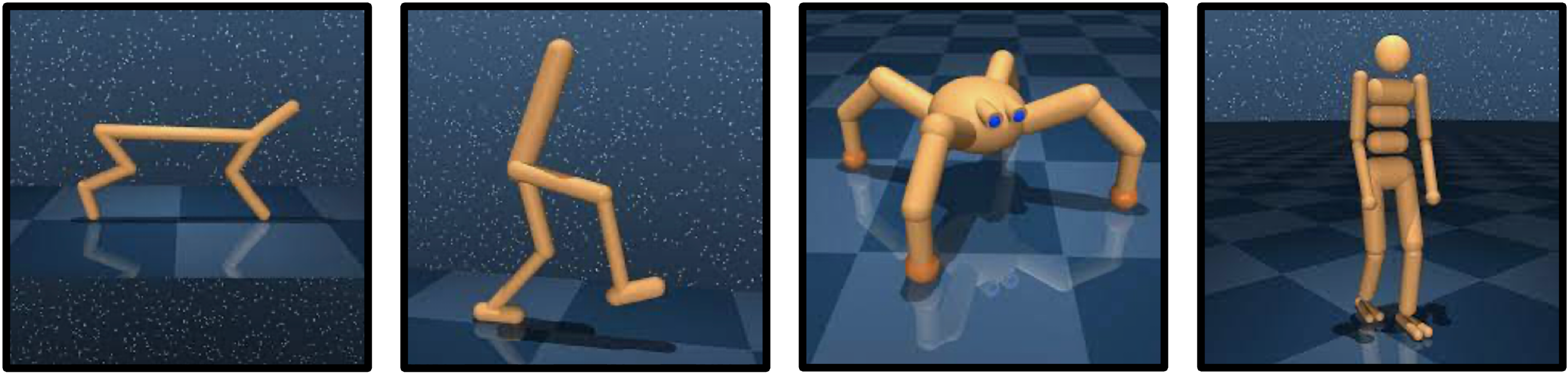}
  \end{center}
  \caption{Continuous control environments in \benchacro.}
  \label{sec3:fig:mood_envs}
\end{wrapfigure}
As introduced in Section~\ref{sec:3evaluation_mood}, \benchname (\benchacro) allows to evaluate offline agents for tasks with increasing levels of complexity, and assesses their ability to make use of additional data coming from behavior policies trained with different objectives. Our benchmark is based on four environments (see Fig.~\ref{sec3:fig:mood_envs}), fifteen tasks, and eighteen base datasets. Each base dataset is collected with a unique objective. We consider either traditional reward maximization objectives for one of the DMC tasks we use for offline training, or the intrinsic objective from Random Network Distillation (RND) \citep{RND}. The only exception is for the humanoid environment, where we only consider the reward maximization objectives. The reason for our choice is that the humanoid model appears quite unstable, with the agent easily losing its balance and collapsing to the ground, a situation from which recovering appears very difficult.
Mainly due to this challenge, we found that optimizing an agent with an RND objective fails to capture almost any meaningful kind of behavior, in stark contrast to the other considered environments.

To obtain each of the base datasets for the considered DMC tasks, we start by collecting replay buffer data from training five TD3 agents using different random seeds for a varying number of steps based on the difficulty of the relative environments. Then, we merge the different whole replay buffers, which we note contain trajectories collected throughout the full training procedure, as we do not impose any size limitation during training. We provide the hyper-parameters of the employed TD3 agents in Table~\ref{appA:tab:hyper_collTD3}.

Hence, as described in Section~\ref{sec:3evaluation_mood}, we then proceed to relabel all the rewards in each base dataset for all other tasks based on the same environment to produce the same, cross, and mixed objective datasets. When evaluating agents on any \benchacro dataset, the reported results represent a normalized percentage, which we simply computed by dividing the collected returns with task-specific targets based on the highest trajectory returns in each of the relative datasets.  We refer to Table~\ref{appA:tab:mood_data_coll} for the task-specific details regarding dataset collection and offline training.

\begin{table}[h]
    
    \label{appA:tab:hyper_collTD3}
    \vspace{10pt}
    \begin{center}
    \begin{tabular}{@{}lc@{}}
    \toprule
    \multicolumn{2}{c}{Online TD3 hyper-parameters}                                                   \\ \midrule
    buffer size $|B|$                           & $\infty$                                  \\
    batch size $|b|$                            & $512$                                     \\
    minimum data to train                       & $5000$                                    \\
    optimizer                                   & Adam                                      \\
    learning rate                               & $0.0003$                                   \\
    policy delay                                & $2$                                       \\
    discount $\gamma$                           & $0.99$                                    \\
    polyak coefficient $\rho$                   & $0.995$                                   \\
    policy/Q network hidden layers              & $2$                                       \\
    policy/Q network hidden dimensionality      & $256$                                     \\
    exploration noise                           & $0.2$                                     \\ \bottomrule
    \end{tabular}
    \caption{Agent hyper-parameters used for data collection in \benchacro.} 
    \end{center}
\end{table}

\begin{table}[H]
\vspace{10pt}
\tabcolsep=0.04cm
\centering
\adjustbox{max width=0.7\linewidth}{
\begin{tabular}{@{}lcccccc@{}}
\toprule
Environment & $|S|$ & $|A|$ & Base dataset objective & Collection steps & Subsampled size & Normalization return target \\ \midrule
            &     &     & walk                   & 5$\times$1M          & 500K                          & 990                           \\
            &     &     & run                    & 5$\times$1M          & 500K                          & 800                           \\
cheetah     & 17  & 6   & walk\_backward         & 5$\times$1M          & 500K                          & 990                           \\
            &     &     & run\_backward          & 5$\times$1M          & 500K                          & 550                           \\
            &     &     & RND                    & 5$\times$2M          & 1M                            & N/A                           \\ \cmidrule(l){2-7} 
            &     &     & walk                   & 5$\times$1M          & 500K                          & 970                           \\
            &     &     & run                    & 5$\times$1M          & 500K                          & 730                           \\
walker      & 24  & 6   & stand                  & 5$\times$1M          & 500K                          & 990                           \\
            &     &     & spin                   & 5$\times$1M          & 500K                          & 990                           \\
            &     &     & RND                    & 5$\times$2M          & 1M                            & N/A                           \\ \cmidrule(l){2-7} 
            &     &     & walk                   & 5$\times$1M          & 500K                          & 940                           \\
            &     &     & run                    & 5$\times$1M          & 500K                          & 800                           \\
quadruped   & 78  & 12  & stand                  & 5$\times$1M          & 500K                          & 970                           \\
            &     &     & jump                   & 5$\times$1M          & 500K                          & 870                           \\
            &     &     & RND                    & 5$\times$2M          & 1M                            & N/A                           \\ \cmidrule(l){2-7} 
            &     &     & walk                   & 5$\times$10M         & 5M                            & 900                           \\
humanoid    & 67  & 21  & run                    & 5$\times$10M         & 5M                            & 400                           \\
            &     &     & stand                  & 5$\times$10M         & 5M                            & 960                           \\ \bottomrule
\end{tabular}
}
\caption{\benchacro datasets collection details.}
\label{appA:tab:mood_data_coll}
\end{table}

\section{Implementation Details}
\label{appB:impl_details}
In this section we provide additional information about our experiments.
\subsection{Network Architecture}
\begin{figure}[t]
    \begin{minipage}{.45\textwidth}
    \begin{center}
    \includegraphics[width=0.5\textwidth]{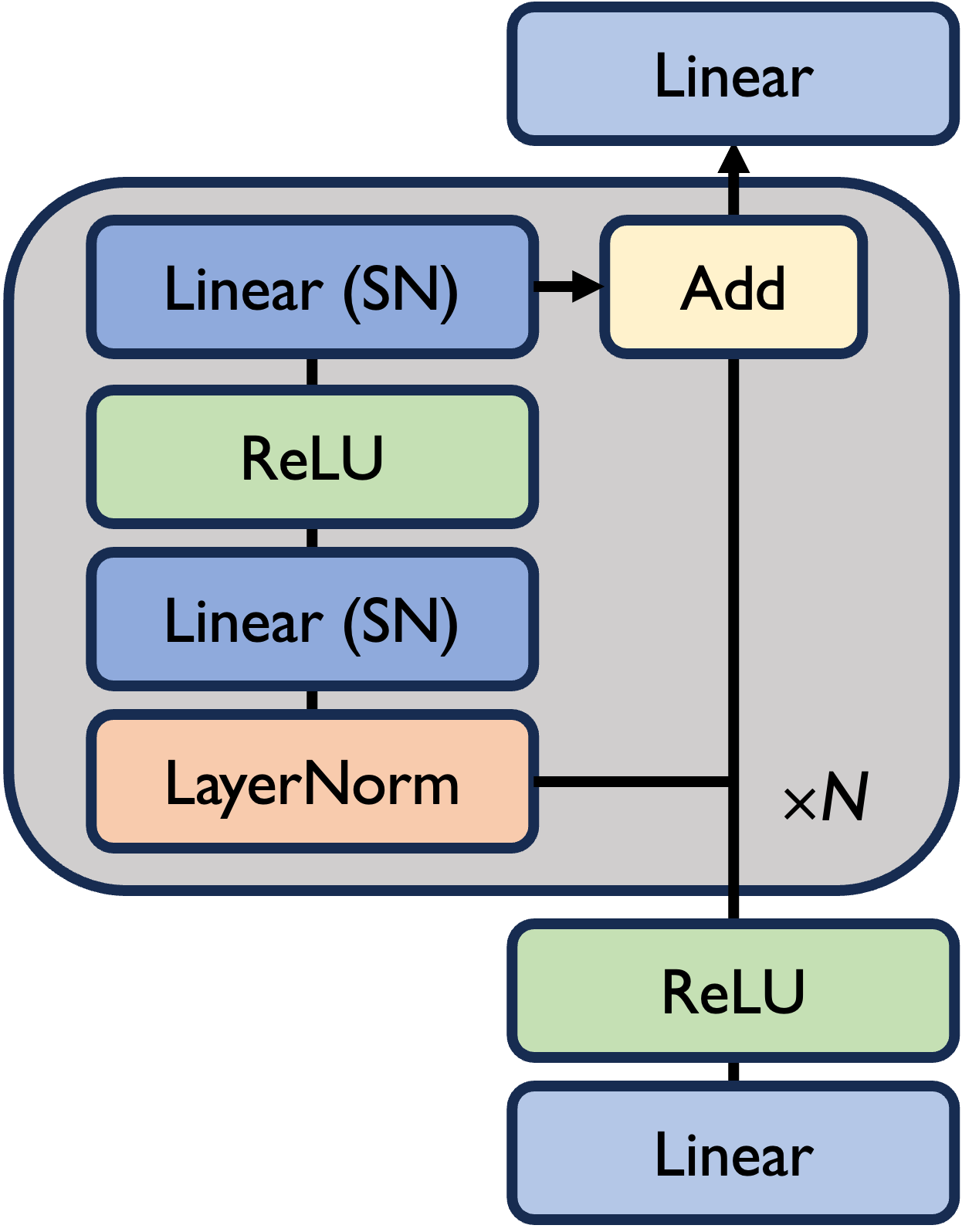}
    \caption{The modern architecture proposed by \citet{deeper-deep-RL} to enable stable training of very deep networks. Each of $N$ residual blocks (gray part) consists of a layer normalization followed by two linear layers regularized via spectral normalization (SN) with a ReLU non-linearity in between.}
    \label{fig:modern}
    \end{center} 
\end{minipage}\hfill
\begin{minipage}{.5\textwidth}
    
    \footnotesize
\centering
\adjustbox{max width=0.98\linewidth}{
    \centering
    \begin{tabular}{|l|c|}
        \toprule
        Testbed  & Alg. Parameters\\
        \midrule
        \midrule
        Common & batch size = 512 \\
        & discount $\gamma$ = $0.99$ \\
        & polyak coefficient $\rho$ = $0.995$ \\
        & M = 50\\
        \midrule
        \midrule
        mood & $\lambda=0.5$, $A_{\mathrm{max}}=\infty$\\
        & train steps (cheetah, walker, hopper) = 1.5M \\
        & train steps (humanoid) = 5M \\
        \midrule
        antmaze-v0 & $\lambda=0$, $A_{\mathrm{max}}=1$\\
        & train steps = 2M \\
        \midrule
        locomotion-0.1-v2 & $\lambda=0.5$, $A_{\mathrm{max}}=\infty$\\
        & train steps = 2M \\
        \midrule
        locomotion-v2 & $\lambda=0.5$, $A_{\mathrm{max}}=\infty$\\
        & train steps = 2M \\
        \midrule
        \bottomrule
    \end{tabular}
    }
    \caption{The parameters common across algorithms, used for each testbed}
    \label{tab:param_per_dataset}
\end{minipage}
  \end{figure}

 In table~\ref{fig:net_size}, we report the activation functions, number of hidden layers, hidden dimension and number of modern blocks we used for small and large networks for each of actor, critic and value networks. Note that the simple and modern networks have equal capacity, disregarding the marginal increase in parameters introduced by layer normalization. Figure~\ref{fig:modern} portrays a block of the modern architecture.

\begin{table*}[h!]
    \footnotesize
\centering
\adjustbox{max width=0.98\linewidth}{
    \begin{tabular}{|l|c|c|c|}
        \toprule
        Components & Simple-Small & Simple-Large & Modern\\
        \midrule
        & activation=ReLu & activation=ReLu& activation=ReLu \\
        \midrule
       Actor& \makecell{hidden layers = 2\\with hidden dim=256}&  \makecell{hidden layers = 5\\with hidden dim=1024} & \makecell{blocks = 2, \\with hidden dim=1024}\\
        \midrule
        Critic&
        \makecell{ hidden layers = 2\\with hidden dim=256}& \makecell{ hidden layers = 3\\with hidden dim=256}& \makecell{ blocks = 1,\\with hidden dim=256}\\
        \midrule
        Value (only IQL)&
        \makecell{ hidden layers = 2\\ with hidden dim=256} & \makecell{hidden layers = 2\\ with hidden dim=1024}& 
        \makecell{simple hidden layers = 2\\ with hidden dim=1024} \\
        \bottomrule
    \end{tabular}
    }
    \caption{Network specification for small, large and modern networks employed by different algorithms}
    \label{fig:net_size}
\end{table*}

\subsection{Algorithmic details}
\textit{Advantage clipping:} Similarly to what done in previous papers we may use advantage clipping in the actor update to avoid numerical overflow due the exponentiation. The actor update is then  
    \begin{gather}
    \label{eq:awac_obj}
        \argmax_\theta\E_{a, s\sim \calB}\left[\frac{\exp(\min\{A_{\phi_1}(s,a), A_{\mathrm{max}}\}/\beta)}{Z}\log \pi_\theta(a|s)\right].
    \end{gather}
    where $Z = \E_{s, a\sim \calB }\left[\exp(\min\{A_{\phi_1}(s,a), A_{\mathrm{max}}\}/\beta)\right]$.

\textit{Pessimism penalty:} As explained in section~\ref{sec:4conjectures} for the ensemble of critics, we employ a hyperparameter $\lambda$ to regulate the extent to which we penalize discrepancies among the critics. We maintain a constant value of $\lambda = 0.5$ across all settings, except in antmaze, where we observe that $\lambda = 0$ (no penalty) is advantageous. This is likely due to the sparsity of rewards in antmaze domains.

Table~\ref{tab:param_per_dataset} highlights the distinct hyperparameters per each testbed.

\subsection{Hyperparameter Sweep} 
The table~\ref{tab:sweep_range} summarizes the range of hyperparameters sweep we used in our experiments for each algorithms and testbed. 

\begin{table*}[h!]
    \footnotesize
\centering
\adjustbox{max width=0.98\linewidth}{
    \centering
\begin{tabular}{|l|c|c|c|c|}
        \toprule
        Testbed/Algorithm & IQL & AWAC & ASAC & TD3+BC\\
        \midrule
        \multirow{ 2}{*}{mood} & \multicolumn{3}{c|}{$\beta = \{0.1, 0.5, 1.5, 3\}$} &  $\alpha = \{0, 0.01,0.1,1,10\}$\\
        \cmidrule(l){2-5} 
        & $\tau = \{0.7, 0.9\}$ & & &\\
        \cmidrule(l){2-5} 
        & \multicolumn{4}{c|}{$\text{learning rate} = \{0.0001, 0.0003\}$}\\
        \midrule
        \midrule
        \multirow{ 2}{*}{antmaze-v0} & \multicolumn{3}{c|}{$\beta = \{0.03, 0.06, 0.1,  0.3 , 0.5\}$} &  $\alpha = \{0.01,  0.1,   1,   10\}$\\
        \cmidrule(l){2-5} 
        & $\tau = \{0.7, 0.9\}$ & & &\\
        \cmidrule(l){2-5} 
        & \multicolumn{4}{c|}{$\text{learning rate} = \{0.0001, 0.0003\}$}\\
        \midrule
        \midrule
        \multirow{ 2}{*}{locomotion-0.1-v2} & \multicolumn{2}{c|}{$\beta = \{0.1,  0.5 , 1.5 , 3 , 10 , 20\}$} & X &  $\alpha = \{0.01,  0.1,   1,   10\}$\\
        \cmidrule(l){2-5} 
        & $\tau = \{0.7, 0.9\}$ & & &\\
        \cmidrule(l){2-5} 
        & \multicolumn{4}{c|}{$\text{learning rate} = \{0.0001, 0.0003\}$}\\
        \midrule
        \midrule
        \multirow{ 2}{*}{locomotion-v2} & $\beta = \{0.1,  0.5 , 1.5 , 3\}$ & \multicolumn{2}{c|}{$\beta = \{0.1, 0.3, 0.5, 1.5, 3\}$} & $\alpha =\{0, 0.01, 0.1, 1, 10\}$\\
        \cmidrule(l){2-5} 
        & $\tau = \{0.7, 0.9\}$ & & &\\
        \cmidrule(l){2-5} 
        & \multicolumn{4}{c|}{$\text{learning rate} = \{0.0001, 0.0003\}$}\\
        \bottomrule
    \end{tabular}
    }
    \caption{Hyperparameter Sweep range per algorithms and per testbed}
    \label{tab:sweep_range}
    \end{table*}

\section{Full Empirical Results}\label{app:all-results}
    \clearpage
    \subsection{MOOD: Full Results}
        \begin{table}[ht]
            \centering
        \begin{minipage}{.45\textwidth}
                \adjustbox{max width=\linewidth}{
                    \begin{tabular}{l|cccc}
    \toprule
    & \multicolumn{4}{c}{mean/stdErr} \\
    agent & IQL & AWAC & TD3 & TD3+BC \\
    actor dim & 256 & 256 & 256 & 256 \\
    value archi dim & 256 & 0 & 0 & 0 \\
    num cri & 2 & 2 & 2 & 2 \\
    critic num layers & 2 & 2 & 2 & 2 \\
    actor num layers & 2 & 2 & 2 & 2 \\
    critic model & simple & simple & simple & simple \\
\bottomrule
    \end{tabular}
    
                }
                \caption{Same Task Dataset: best scores with small architecture}
        \end{minipage}\hfill
        \begin{minipage}{.45\textwidth}
                \adjustbox{max width=\linewidth}{
                    \begin{tabular}{l|cccc}
    \toprule
    & \multicolumn{4}{c}{mean/stdErr} \\
    agent & IQL & AWAC & TD3 & TD3+BC \\
    actor dim & 256 & 256 & 256 & 256 \\
    value archi dim & 256 & 0 & 0 & 0 \\
    num cri & 2 & 2 & 2 & 2 \\
    critic num layers & 2 & 2 & 2 & 2 \\
    actor num layers & 2 & 2 & 2 & 2 \\
    critic model & simple & simple & simple & simple \\
\bottomrule
    \end{tabular}
    
                }
                \caption{Mixed Task Dataset: best scores with small architecture}
        \end{minipage}
    \end{table}
    \begin{table}[ht]
        \centering
    \begin{minipage}{.45\textwidth}
            \adjustbox{max width=\linewidth}{
                \begin{tabular}{llllll}
    \toprule
     & \multicolumn{5}{r}{mean/sem} \\
    agent & iql2 & td3as & td3aw & td3p & td3p+bc \\
    actor dim & 1024 & 1024 & 1024 & 1024 & 1024 \\
    value archi dim & 1024 & 0 & 0 & 0 & 0 \\
    num cri & 2 & 2 & 2 & 2 & 2 \\
    critic num layers & 3 & 3 & 3 & 3 & 3 \\
    actor num layers & 5 & 5 & 5 & 5 & 5 \\
    critic model & simple & simple & simple & simple & simple \\
\bottomrule
    \end{tabular}
    
            }
            \caption{Mixed Task Dataset: best scores with large architecture and 2 critics}
    \end{minipage}\hfill
    \begin{minipage}{.45\textwidth}
            \adjustbox{max width=\linewidth}{
                \begin{tabular}{llllll}
    \toprule
     & \multicolumn{5}{r}{mean/sem} \\
    agent & iql2 & td3as & td3aw & td3p & td3p+bc \\
    actor dim & 1024 & 1024 & 1024 & 1024 & 1024 \\
    value archi dim & 1024 & 0 & 0 & 0 & 0 \\
    num cri & 2 & 2 & 2 & 2 & 2 \\
    critic num layers & 2 & 2 & 2 & 2 & 2 \\
    actor num layers & 2 & 2 & 2 & 2 & 2 \\
    critic model & modern & modern & modern & modern & modern \\
\bottomrule
    \end{tabular}
    
            }
            \caption{Mixed Task Dataset: best scores with modern architecture and 2 critics}
    \end{minipage}
\end{table}

    \begin{table}[ht]
        \adjustbox{max width=0.99\linewidth}{
            \centering
            \begin{tabular}{lcccccccccc}
    \toprule
     & \multicolumn{10}{c}{mean/sem} \\
    agent & \multicolumn{2}{c}{IQL} & \multicolumn{2}{c}{ASAC} & \multicolumn{2}{c}{AWAC} & \multicolumn{2}{c}{TD3} & \multicolumn{2}{c}{TD3+BC} \\
    actor dim & \multicolumn{2}{c}{1024} & \multicolumn{2}{c}{1024} & \multicolumn{2}{c}{1024} & \multicolumn{2}{c}{1024} & \multicolumn{2}{c}{1024} \\
    value archi dim & \multicolumn{2}{c}{1024} & \multicolumn{2}{c}{0} & \multicolumn{2}{c}{0} & \multicolumn{2}{c}{0} & \multicolumn{2}{c}{0} \\
    num cri & \multicolumn{2}{c}{5} & \multicolumn{2}{c}{5} & \multicolumn{2}{c}{5} & \multicolumn{2}{c}{5} & \multicolumn{2}{c}{5} \\
    critic num layers & \multicolumn{2}{c}{3} & \multicolumn{2}{c}{3} & \multicolumn{2}{c}{3} & \multicolumn{2}{c}{3} & \multicolumn{2}{c}{3} \\
    actor num layers & \multicolumn{2}{c}{5} & \multicolumn{2}{c}{5} & \multicolumn{2}{c}{5} & \multicolumn{2}{c}{5} & \multicolumn{2}{c}{5} \\
    critic model & \multicolumn{2}{c}{simple} & \multicolumn{2}{c}{simple} & \multicolumn{2}{c}{simple} & \multicolumn{2}{c}{simple} & \multicolumn{2}{c}{simple} \\
\bottomrule
    \end{tabular}

        }
        \caption{Mixed Task Dataset: best scores with large architecture and 5 critics}
        \vspace{.2in}
        \adjustbox{max width=0.99\linewidth}{
            \centering
            \begin{tabular}{lcccccccccc}
    \toprule
     & \multicolumn{10}{c}{mean/sem} \\
     agent & \multicolumn{2}{c}{IQL} & \multicolumn{2}{c}{ASAC} & \multicolumn{2}{c}{AWAC} & \multicolumn{2}{c}{TD3} & \multicolumn{2}{c}{TD3+BC} \\
    actor dim & \multicolumn{2}{c}{1024} & \multicolumn{2}{c}{1024} & \multicolumn{2}{c}{1024} & \multicolumn{2}{c}{1024} & \multicolumn{2}{c}{1024} \\
    value archi dim & \multicolumn{2}{c}{1024} & \multicolumn{2}{c}{0} & \multicolumn{2}{c}{0} & \multicolumn{2}{c}{0} & \multicolumn{2}{c}{0} \\
    num cri & \multicolumn{2}{c}{5} & \multicolumn{2}{c}{5} & \multicolumn{2}{c}{5} & \multicolumn{2}{c}{5} & \multicolumn{2}{c}{5} \\
    critic num layers & \multicolumn{2}{c}{2} & \multicolumn{2}{c}{2} & \multicolumn{2}{c}{2} & \multicolumn{2}{c}{2} & \multicolumn{2}{c}{2} \\
    actor num layers & \multicolumn{2}{c}{2} & \multicolumn{2}{c}{2} & \multicolumn{2}{c}{2} & \multicolumn{2}{c}{2} & \multicolumn{2}{c}{2} \\
    critic model & \multicolumn{2}{c}{modern} & \multicolumn{2}{c}{modern} & \multicolumn{2}{c}{modern} & \multicolumn{2}{c}{modern} & \multicolumn{2}{c}{modern} \\
\bottomrule
    \end{tabular}
    
        }
        \caption{Mixed Task Dataset: best scores with modern architecture and 5 critics}

    \end{table}

    \clearpage
    \subsection{Locomotion-v2: Full Results}
        \begin{table}[ht]
            \adjustbox{max width=0.60\linewidth}{
                \centering
                \begin{tabular}{lcccc}
    \toprule
     & \multicolumn{4}{c}{mean/sem} \\
    agent & IQL & AWAC & TD3 & TD3+BC \\
    actor dim & 256 & 256 & 256 & 256 \\
    num cri & 2 & 2 & 2 & 2 \\
    critic num layers & 2 & 2 & 2 & 2 \\
    actor num layers & 2 & 2 & 2 & 2 \\
    critic model & simple & simple & simple & simple \\
    actor model & simple & simple & simple & simple \\
    num es & 1 & 1 & 1 & 1 \\
    task &  &  &  &  \\
    \midrule
 halfcheetah-medium-expert-v2 & 87.6 / 1.2 & 93.0 / 0.5 & 3.9 / nan & 91.3 / 0.4 \\
 halfcheetah-medium-replay-v2 & 45.0 / 0.2 & 48.0 / 0.2 & 42.0 / 2.8 & 47.6 / 0.2 \\
 halfcheetah-medium-v2 & 51.6 / 0.1 & 51.9 / 0.2 & 44.9 / 7.4 & 66.2 / 0.2 \\
 hopper-medium-expert-v2 & 107.9 / 0.6 & 100.8 / 1.8 & 0.7 / 0.0 & 95.5 / 4.0 \\
 hopper-medium-replay-v2 & 96.7 / 0.7 & 99.1 / 1.0 & 64.8 / 21.3 & 79.8 / 4.3 \\
 hopper-medium-v2 & 97.4 / 3.2 & 79.6 / 1.6 & NaN & 72.9 / 18.1 \\
 walker2d-medium-expert-v2 & 112.2 / 0.9 & 109.7 / 0.1 & -0.1 / 0.0 & 108.9 / 2.7 \\
 walker2d-medium-replay-v2 & 88.0 / 1.5 & 82.6 / 1.7 & 3.5 / 0.6 & 75.7 / 5.1 \\
 walker2d-medium-v2 & 85.4 / 1.3 & 83.5 / 2.4 & -0.1 / 0.1 & 80.7 / 0.1 \\
 \midrule
    total & 771.8 & 748.2 & 159.7 & 718.6 \\
    \bottomrule
    \end{tabular}
    
            }
            \caption{Locomotion-v2: best scores with small architecture}
        \end{table}
            \begin{table}[t]
            \centering
            \adjustbox{max width=0.99\linewidth}{
                \centering
                \tabcolsep=0.06cm
                \begin{tabular}{lcccccccccc}
    \toprule
     & \multicolumn{10}{c}{mean/sem} \\
     agent & \multicolumn{2}{c}{IQL} & \multicolumn{2}{c}{ASAC} & \multicolumn{2}{c}{AWAC} & \multicolumn{2}{c}{TD3} & \multicolumn{2}{c}{TD3+BC} \\
    actor dim & \multicolumn{2}{c}{1024} & \multicolumn{2}{c}{1024} & \multicolumn{2}{c}{1024} & \multicolumn{2}{c}{1024} & \multicolumn{2}{c}{1024} \\
    num cri & \multicolumn{2}{c}{5} & \multicolumn{2}{c}{5} & \multicolumn{2}{c}{5} & \multicolumn{2}{c}{5} & \multicolumn{2}{c}{5} \\
    critic num layers & \multicolumn{2}{c}{3} & \multicolumn{2}{c}{3} & \multicolumn{2}{c}{3} & \multicolumn{2}{c}{3} & \multicolumn{2}{c}{3} \\
    actor num layers & \multicolumn{2}{c}{5} & \multicolumn{2}{c}{5} & \multicolumn{2}{c}{5} & \multicolumn{2}{c}{5} & \multicolumn{2}{c}{5} \\
    critic model & \multicolumn{2}{c}{simple} & \multicolumn{2}{c}{simple} & \multicolumn{2}{c}{simple} & \multicolumn{2}{c}{simple} & \multicolumn{2}{c}{simple} \\
    actor model & \multicolumn{2}{c}{simple} & \multicolumn{2}{c}{simple} & \multicolumn{2}{c}{simple} & \multicolumn{2}{c}{simple} & \multicolumn{2}{c}{simple} \\
\bottomrule
    \end{tabular}
    
            }
            \caption{Locomotion-v2: best scores with large architecture}
            \vspace{.1in}
            
            \adjustbox{max width=0.99\linewidth}{
                \centering
                \tabcolsep=0.06cm
                \begin{tabular}{lcccccccccc}
    \toprule
     & \multicolumn{10}{c}{mean/sem} \\
     agent & \multicolumn{2}{c}{IQL} & \multicolumn{2}{c}{ASAC} & \multicolumn{2}{c}{AWAC} & \multicolumn{2}{c}{TD3} & \multicolumn{2}{c}{TD3+BC} \\
    actor dim & \multicolumn{2}{c}{1024} & \multicolumn{2}{c}{1024} & \multicolumn{2}{c}{1024} & \multicolumn{2}{c}{1024} & \multicolumn{2}{c}{1024} \\
    num cri & \multicolumn{2}{c}{5} & \multicolumn{2}{c}{5} & \multicolumn{2}{c}{5} & \multicolumn{2}{c}{5} & \multicolumn{2}{c}{5} \\
    critic num layers & \multicolumn{2}{c}{2} & \multicolumn{2}{c}{2} & \multicolumn{2}{c}{2} & \multicolumn{2}{c}{2} & \multicolumn{2}{c}{2} \\
    actor num layers & \multicolumn{2}{c}{2} & \multicolumn{2}{c}{2} & \multicolumn{2}{c}{2} & \multicolumn{2}{c}{2} & \multicolumn{2}{c}{2} \\
    critic model & \multicolumn{2}{c}{modern} & \multicolumn{2}{c}{modern} & \multicolumn{2}{c}{modern} & \multicolumn{2}{c}{modern} & \multicolumn{2}{c}{modern} \\
    actor model & \multicolumn{2}{c}{modern} & \multicolumn{2}{c}{modern} & \multicolumn{2}{c}{modern} & \multicolumn{2}{c}{modern} & \multicolumn{2}{c}{modern} \\
\bottomrule
    \end{tabular}
    
            }
            \caption{Locomotion-v2: best scores with modern architecture}
        \end{table}

\clearpage
\subsection{Antmaze-v0: Full Results}
    \begin{table}[ht]
        \centering
        \adjustbox{max width=0.7\linewidth}{
            \centering
            \begin{tabular}{l|ccc}
    \toprule
     & \multicolumn{3}{c}{mean/sem} \\
    agent & IQL & AWAC & TD3+BC \\
    actor dim & 256 & 256 & 256 \\
    num cri & 2 & 2 & 2 \\
    critic num layers & 2 & 2 & 2 \\
    actor num layers & 2 & 2 & 2 \\
    critic model & simple & simple & simple \\
    actor model & simple & simple & simple \\
    score processing & softmax & softmax & none \\
    num es & 1 & 1 & 1 \\
    task &  &  &  \\
    \midrule
     antmaze-large-diverse-v0 & 41.5 / 3.4 & 36.5 / 2.2 & 2.1 / 1.7 \\
     antmaze-large-play-v0 & 28.7 / 9.0 & 7.8 / 7.8 & 0.2 / 0.2 \\
     antmaze-medium-diverse-v0 & 60.7 / 7.9 & 31.4 / 19.3 & 43.9 / 3.0 \\
     antmaze-medium-play-v0 & 66.3 / 2.1 & 76.3 / 1.6 & 28.1 / 11.7 \\
     antmaze-umaze-diverse-v0 & 68.8 / 7.3 & 14.0 / 14.0 & 41.4 / 3.3 \\
     antmaze-umaze-v0 & 89.1 / 3.4 & 45.6 / 15.0 & 70.2 / 15.8 \\
     \midrule
    total & 355.2 & 211.6 & 186.0 \\
    \bottomrule
    \end{tabular}
    
        }
        \caption{Antmaze-v0: best scores with small architecture}
    \end{table}
        \vspace{.2in}

    \begin{table}[ht]
        \adjustbox{max width=0.99\linewidth}{
            \centering
            \begin{tabular}{lcccccccc}
    \toprule
     & \multicolumn{8}{c}{mean/sem} \\
    agent & \multicolumn{2}{c}{IQL} & \multicolumn{2}{c}{ASAC} & \multicolumn{2}{c}{AWAC} & \multicolumn{2}{c}{TD3+BC} \\
    actor dim & \multicolumn{2}{c}{1024} & \multicolumn{2}{c}{1024} & \multicolumn{2}{c}{1024} & \multicolumn{2}{c}{1024} \\
    num cri & \multicolumn{2}{c}{5} & \multicolumn{2}{c}{5} & \multicolumn{2}{c}{5} & \multicolumn{2}{c}{5} \\
    critic num layers & \multicolumn{2}{c}{3} & \multicolumn{2}{c}{3} & \multicolumn{2}{c}{3} & \multicolumn{2}{c}{3} \\
    actor num layers & \multicolumn{2}{c}{5} & \multicolumn{2}{c}{5} & \multicolumn{2}{c}{5} & \multicolumn{2}{c}{5} \\
    critic model & \multicolumn{2}{c}{simple} & \multicolumn{2}{c}{simple} & \multicolumn{2}{c}{simple} & \multicolumn{2}{c}{simple} \\
    actor model & \multicolumn{2}{c}{simple} & \multicolumn{2}{c}{simple} & \multicolumn{2}{c}{simple} & \multicolumn{2}{c}{simple} \\
\bottomrule
    \end{tabular}
    
        }
        \caption{Antmaze-v0: best scores with 5 critics and large architecture}
        \vspace{.2in}
        \adjustbox{max width=0.99\linewidth}{
            \centering
            \begin{tabular}{lcccccccc}
    \toprule
     & \multicolumn{8}{c}{mean/sem} \\
     agent & \multicolumn{2}{c}{IQL} & \multicolumn{2}{c}{ASAC} & \multicolumn{2}{c}{AWAC} & \multicolumn{2}{c}{TD3+BC} \\
    actor dim & \multicolumn{2}{c}{1024} & \multicolumn{2}{c}{1024} & \multicolumn{2}{c}{1024} & \multicolumn{2}{c}{1024} \\
    num cri & \multicolumn{2}{c}{5} & \multicolumn{2}{c}{5} & \multicolumn{2}{c}{5} & \multicolumn{2}{c}{5} \\
    critic num layers & \multicolumn{2}{c}{2} & \multicolumn{2}{c}{2} & \multicolumn{2}{c}{2} & \multicolumn{2}{c}{2} \\
    actor num layers & \multicolumn{2}{c}{2} & \multicolumn{2}{c}{2} & \multicolumn{2}{c}{2} & \multicolumn{2}{c}{2} \\
    critic model & \multicolumn{2}{c}{modern} & \multicolumn{2}{c}{modern} & \multicolumn{2}{c}{modern} & \multicolumn{2}{c}{modern} \\
    actor model & \multicolumn{2}{c}{modern} & \multicolumn{2}{c}{modern} & \multicolumn{2}{c}{modern} & \multicolumn{2}{c}{modern} \\
\bottomrule
    \end{tabular}
    
        }
        \caption{Antmaze-v0: best scores with 5 critics and modern architecture}
    \end{table}
        \begin{table}[ht]

        \adjustbox{max width=0.99\linewidth}{
            \centering
            \begin{tabular}{lcccccccc}
    \toprule
     & \multicolumn{8}{c}{mean/sem} \\
     agent & \multicolumn{2}{c}{IQL} & \multicolumn{2}{c}{ASAC} & \multicolumn{2}{c}{AWAC} & \multicolumn{2}{c}{TD3+BC} \\
    actor dim & \multicolumn{2}{c}{1024} & \multicolumn{2}{c}{1024} & \multicolumn{2}{c}{1024} & \multicolumn{2}{c}{1024} \\
    num cri & \multicolumn{2}{c}{10} & \multicolumn{2}{c}{10} & \multicolumn{2}{c}{10} & \multicolumn{2}{c}{10} \\
    critic num layers & \multicolumn{2}{c}{3} & \multicolumn{2}{c}{3} & \multicolumn{2}{c}{3} & \multicolumn{2}{c}{3} \\
    actor num layers & \multicolumn{2}{c}{5} & \multicolumn{2}{c}{5} & \multicolumn{2}{c}{5} & \multicolumn{2}{c}{5} \\
    critic model & \multicolumn{2}{c}{simple} & \multicolumn{2}{c}{simple} & \multicolumn{2}{c}{simple} & \multicolumn{2}{c}{simple} \\
    actor model & \multicolumn{2}{c}{simple} & \multicolumn{2}{c}{simple} & \multicolumn{2}{c}{simple} & \multicolumn{2}{c}{simple} \\
\bottomrule
    \end{tabular}
    
        }
        \caption{Antmaze-v0: best scores with 10 critics and large architecture}
        \vspace{.2in}
        \adjustbox{max width=0.99\linewidth}{
            \centering
            \begin{tabular}{lcccccccc}
    \toprule
     & \multicolumn{8}{c}{mean/sem} \\
     agent & \multicolumn{2}{c}{IQL} & \multicolumn{2}{c}{ASAC} & \multicolumn{2}{c}{AWAC} & \multicolumn{2}{c}{TD3+BC} \\
    actor dim & \multicolumn{2}{c}{1024} & \multicolumn{2}{c}{1024} & \multicolumn{2}{c}{1024} & \multicolumn{2}{c}{1024} \\
    num cri & \multicolumn{2}{c}{10} & \multicolumn{2}{c}{10} & \multicolumn{2}{c}{10} & \multicolumn{2}{c}{10} \\
    critic num layers & \multicolumn{2}{c}{2} & \multicolumn{2}{c}{2} & \multicolumn{2}{c}{2} & \multicolumn{2}{c}{2} \\
    actor num layers & \multicolumn{2}{c}{2} & \multicolumn{2}{c}{2} & \multicolumn{2}{c}{2} & \multicolumn{2}{c}{2} \\
    critic model & \multicolumn{2}{c}{modern} & \multicolumn{2}{c}{modern} & \multicolumn{2}{c}{modern} & \multicolumn{2}{c}{modern} \\
    actor model & \multicolumn{2}{c}{modern} & \multicolumn{2}{c}{modern} & \multicolumn{2}{c}{modern} & \multicolumn{2}{c}{modern} \\
\bottomrule
    \end{tabular}
    
        }
        \caption{Antmaze-v0: best scores with 10 critics and modern architecture}
    \end{table}

    \clearpage
    \subsection{Locomotion-0.1-v2: Full Results}
    \begin{table}[ht]

    \adjustbox{max width=0.99\linewidth}{
        \centering
        \begin{tabular}{lllllllll}
    \toprule
     & \multicolumn{8}{c}{mean/sem} \\
     agent & \multicolumn{2}{c}{IQL} & \multicolumn{2}{c}{ASAC} & \multicolumn{2}{c}{AWAC} & \multicolumn{2}{c}{TD3+BC} \\
    actor dim & \multicolumn{2}{c}{1024} & \multicolumn{2}{c}{1024} & \multicolumn{2}{c}{1024} & \multicolumn{2}{c}{1024} \\
    num cri & \multicolumn{2}{c}{10} & \multicolumn{2}{c}{10} & \multicolumn{2}{c}{10} & \multicolumn{2}{c}{10} \\
    critic num layers & \multicolumn{2}{c}{3} & \multicolumn{2}{c}{3} & \multicolumn{2}{c}{3} & \multicolumn{2}{c}{3} \\
    actor num layers & \multicolumn{2}{c}{5} & \multicolumn{2}{c}{5} & \multicolumn{2}{c}{5} & \multicolumn{2}{c}{5} \\
    critic model & \multicolumn{2}{c}{simple} & \multicolumn{2}{c}{simple} & \multicolumn{2}{c}{simple} & \multicolumn{2}{c}{simple} \\
    actor model & \multicolumn{2}{c}{simple} & \multicolumn{2}{c}{simple} & \multicolumn{2}{c}{simple} & \multicolumn{2}{c}{simple} \\
\bottomrule
\end{tabular}
    
    }
    \caption{Locomotion-0.1-v2: best scores with 10 critics and large architecture}
    \vspace{.2in}
    \adjustbox{max width=0.99\linewidth}{
        \centering
        \begin{tabular}{lllllllll}
    \toprule
     & \multicolumn{8}{c}{mean/sem} \\
     agent & \multicolumn{2}{c}{IQL} & \multicolumn{2}{c}{ASAC} & \multicolumn{2}{c}{AWAC} & \multicolumn{2}{c}{TD3+BC} \\
    actor dim & \multicolumn{2}{c}{1024} & \multicolumn{2}{c}{1024} & \multicolumn{2}{c}{1024} & \multicolumn{2}{c}{1024} \\
    num cri & \multicolumn{2}{c}{10} & \multicolumn{2}{c}{10} & \multicolumn{2}{c}{10} & \multicolumn{2}{c}{10} \\
    critic num layers & \multicolumn{2}{c}{2} & \multicolumn{2}{c}{2} & \multicolumn{2}{c}{2} & \multicolumn{2}{c}{2} \\
    actor num layers & \multicolumn{2}{c}{2} & \multicolumn{2}{c}{2} & \multicolumn{2}{c}{2} & \multicolumn{2}{c}{2} \\
    critic model & \multicolumn{2}{c}{modern} & \multicolumn{2}{c}{modern} & \multicolumn{2}{c}{modern} & \multicolumn{2}{c}{modern} \\
    actor model & \multicolumn{2}{c}{modern} & \multicolumn{2}{c}{modern} & \multicolumn{2}{c}{modern} & \multicolumn{2}{c}{modern} \\
\bottomrule
\end{tabular}
    
    }
    \caption{Locomotion-0.1-v2: best scores with 10 critics and modern architecture}
\end{table}

    \subsection{ExoRL: Full Results}\label{app:exorl}
    We also tested our hypotheses on the ExoRL dataset generated using RND~\citep{yarats2022don}. 
    \begin{table}[ht]
    \centering
    \adjustbox{max width=0.45\linewidth}{

        \begin{tabular}{lllll}
\toprule
{} & \multicolumn{4}{l}{mean/sem} \\
agent &              IQL &             AWAC &              TD3 &           TD3+BC \\
actor dim &               256 &               256 &               256 &               256 \\
num cri &                 2 &                 2 &                 2 &                 2 \\
critic num layers &                 2 &                 2 &                 2 &                 2 \\
actor num layers &                 2 &                 2 &                 2 &                 2 \\
critic model &            simple &            simple &            simple &            simple \\
actor model &            simple &            simple &            simple &            simple \\
\bottomrule
\end{tabular}

    }
    \caption{Exorl: best scores with small network}
    \vspace{.2in}
    \adjustbox{max width=0.99\linewidth}{
        \centering
        \begin{tabular}{lllllllll}
\toprule
{} & \multicolumn{8}{l}{mean/sem} \\
agent & \multicolumn{2}{l}{IQL} & \multicolumn{2}{l}{AWAC} & \multicolumn{2}{l}{TD3} & \multicolumn{2}{l}{TD3+BC} \\
actor  dim & \multicolumn{2}{l}{1024} & \multicolumn{2}{l}{1024} & \multicolumn{2}{l}{1024} & \multicolumn{2}{l}{1024} \\
num  cri & \multicolumn{2}{l}{10} & \multicolumn{2}{l}{10} & \multicolumn{2}{l}{10} & \multicolumn{2}{l}{10} \\
critic  num  layers & \multicolumn{2}{l}{3} & \multicolumn{2}{l}{3} & \multicolumn{2}{l}{3} & \multicolumn{2}{l}{3} \\
actor  num  layers & \multicolumn{2}{l}{5} & \multicolumn{2}{l}{5} & \multicolumn{2}{l}{5} & \multicolumn{2}{l}{5} \\
critic  model & \multicolumn{2}{l}{simple} & \multicolumn{2}{l}{simple} & \multicolumn{2}{l}{simple} & \multicolumn{2}{l}{simple} \\
actor  model & \multicolumn{2}{l}{simple} & \multicolumn{2}{l}{simple} & \multicolumn{2}{l}{simple} & \multicolumn{2}{l}{simple} \\
\bottomrule
\end{tabular}

    }
    \caption{Exorl: best scores with 10 critics and large simple architecture}
   \end{table}

    \clearpage

\newpage
\begin{figure}
\begin{minipage}[t]{0.46\textwidth}
\vspace{0pt}

\begin{algorithm}[H]
    \footnotesize
    \caption{Logsumexp-tree updating}
    \label{app:alg:logsumexp_up}
    \begin{algorithmic}
    \STATE {\textbf{input} leaf node $n$}
    \STATE {\textbf{input} unnormalized log probability $l$}
    \STATE $n.\text{value}\gets l$
    \WHILE {\textbf{not} $n=\text{root}$}
    \STATE $n\gets n.\text{parent}$
    \STATE $a\gets n.\text{leftchild.value}$
    \STATE $b\gets n.\text{rightchild.value}$
    \STATE $m\gets \text{max}(a, b)$
    \STATE $n.\text{value}\gets m+\text{log1p}(e^{a-m} + e^{b-m})$ 
    \ENDWHILE
    \end{algorithmic}
\end{algorithm}

\end{minipage}
\hfill
\begin{minipage}[t]{0.53\textwidth}
\vspace{0pt}

\begin{algorithm}[H]
    \footnotesize
    \caption{Logsumexp-tree sampling}
    \label{app:alg:logsumexp_sampling}
    \begin{algorithmic}
    \STATE Sample $u\sim U[0, 1]$, $r\gets \log{u} + \text{root.value}$, $n\gets \text{root}$
    \WHILE{\textbf{not} leaf($n$)}
    \STATE $v\gets n.\text{leftchild.value}$
    \IF{if $r<v$}
    \STATE $n\gets n.\text{leftchild}$
    \ELSE
    \STATE $n\gets n.\text{rightchild}$, $r\gets r+\text{log1p}(-e^{v-r})$ 
    \ENDIF
    \ENDWHILE
    \STATE \textbf{return} n
    \end{algorithmic}
\end{algorithm}
\end{minipage}
\end{figure}

\section{Logsumexp tree data-structure}

\label{appG:logsumexp-tree}

Sampling according to the action-maximizing constrained distribution given by $\calB^*(s,a)\deq \frac{1}{Z} \calB(s,a) \exp(A_\phi(s,a)/\beta)$ in an efficient and stable manner introduced several challenges, which our new \textit{logsumexp-tree} was designed to address. 

\textbf{Sum-tree summary.} Sampling with unnormalized probabilities can be achieve with a traditional sum-tree. A sum-tree is a data-structure taking the form of a binary tree where the value of each of its nodes corresponds to the value of its children. Hence, we can store unnormalized probabilities in each of $N$ leaf nodes, making the root correspond to the normalizing factor $Z$. Hence, every time an unnormalized probability is updated, we only require $O(\log N)$ iterative updates to recompute the values its ancestors, leaving the other nodes unmodified. Similarly, we can sample from the true distribution via sampling a uniform $r\sim U[0, Z]$ and do $O(\log N)$ comparisons until we reach one of the leaves. In particular, starting from the root, we compare $r$ with a node's left child value $v$: if $v<r$ we descend to the left subtree, otherwise we descend to the right subtree and update $r\gets r-v$. \citet{per} slightly modify the first step of this procedure when sampling an $n$-sized minibatch, by dividing $Z$ into $n$ equal-length segments and obtaining each $r_i\in \{r_{1:n}\}$ by sampling from $U[Z/n\times(i-1), Z/n\times i]$. This is done to collect more `spread-out' samples across each minibatch, something that we found did not seem to play a significant effect on bias or performance.

\textbf{Logsumexp-tree.} In our use case, the unnormalized probablities given by $\calB$ (Equation~\ref{sec4:eqn:prioritized_B_dist}) are the result of a scaled exponentiation whose magnitude appears to notably vary across problem setting and training stage. Hence, in practice, we found that directly recording $\calB^*(s,a)$ into a sum-tree resulted in arithmetic underflow (with many of the leaves and their sums collapsing to zeros) and overflow (leading to crashes due to exceeding the maximum representable values). To address these challenges, the 
logsumexp-tree allows to record the unnormalized logits before exponentiation $q(s,a)=A_\phi(s,a)/\beta$. Moreover, it allows to perform updating and sampling operations with the same $O(\log n)$ complexity as a sum-tree with stable operations without having to store any explicit values outside log space. In particular, each node $p$ in our new data stracture stores the 'logsumexp' of its children $a, b$, an operation that can be stably done via first shifting $a$ and $b$ by their maximum and using the highly-precise log1p operation implemented in Numpy/Pytorch. In a similar fashion, we can now sample by transforming a uniform variable and applying a log transformation. Hence, analogously to the sum-tree, starting with the root node, we can descend the logsumexp-tree by comparing our sample $r\in(-\infty, \log(Z)]$ with its left child's value $v$ to choose which branch to follow. However, this time, if $r\geq v$ we perform a 'logsubstractexp' operation to update the value of $r$. We refer to Algorithms~\ref{app:alg:logsumexp_up} and \ref{app:alg:logsumexp_sampling} for further details and the exact mathematical operations involved. In our shared code, we provide an implementation of the logsumexp-tree stored in a simple array representation, allowing for efficient fully-parallelized operations.

\newpage
\section{The Role of Mean-Seeking}

\begin{figure*}[t]
    \centering
    \includegraphics[width=0.99\linewidth]{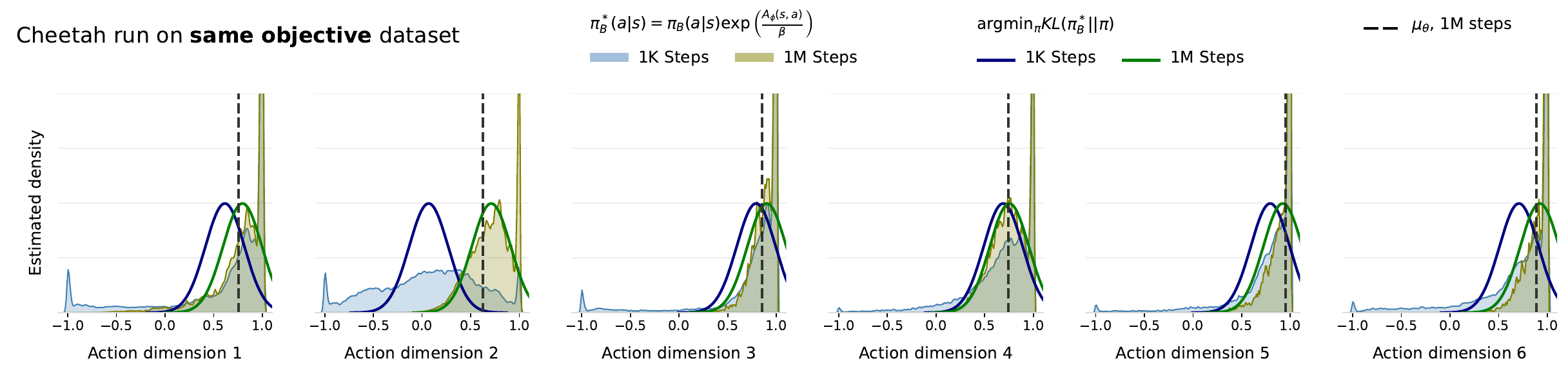}
    \includegraphics[width=0.99\linewidth]
    {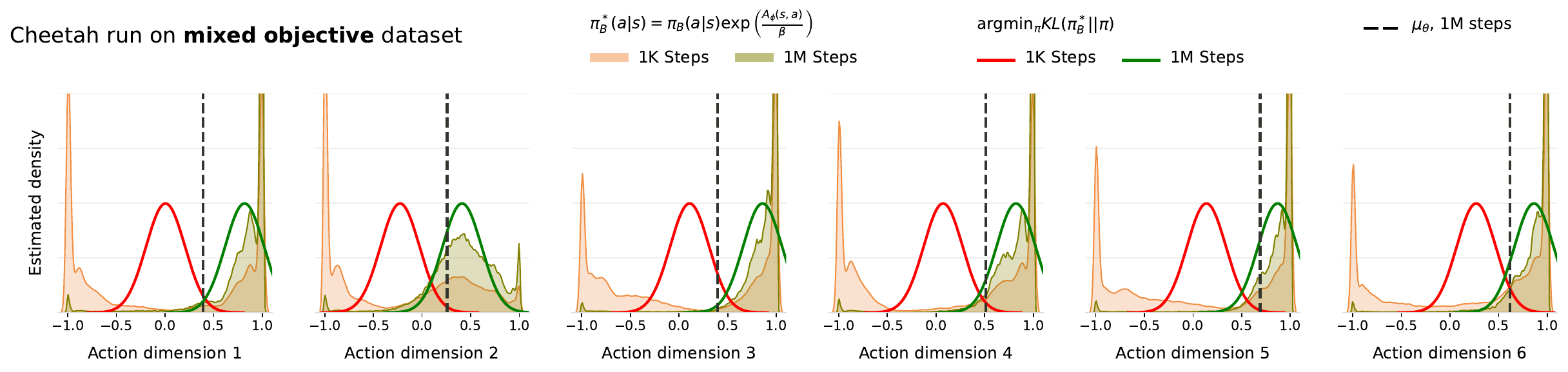}
    \caption{Estimated advantage-weighted action-distribution from $\pi_\calB^*$ and the corresponding optimal projection with a Gaussian $\pi$ after 1K and 1M optimization steps of offline training with AWAC on the cheetah run task with the same objective dataset (A) and the mixed objective dataset (B).}
    \label{appE:fig:dist_evo_st_mt}
  \end{figure*}%

\label{appE:mean_seeking}

\begin{figure}[t]
    \centering
    \includegraphics[width=0.95\linewidth]{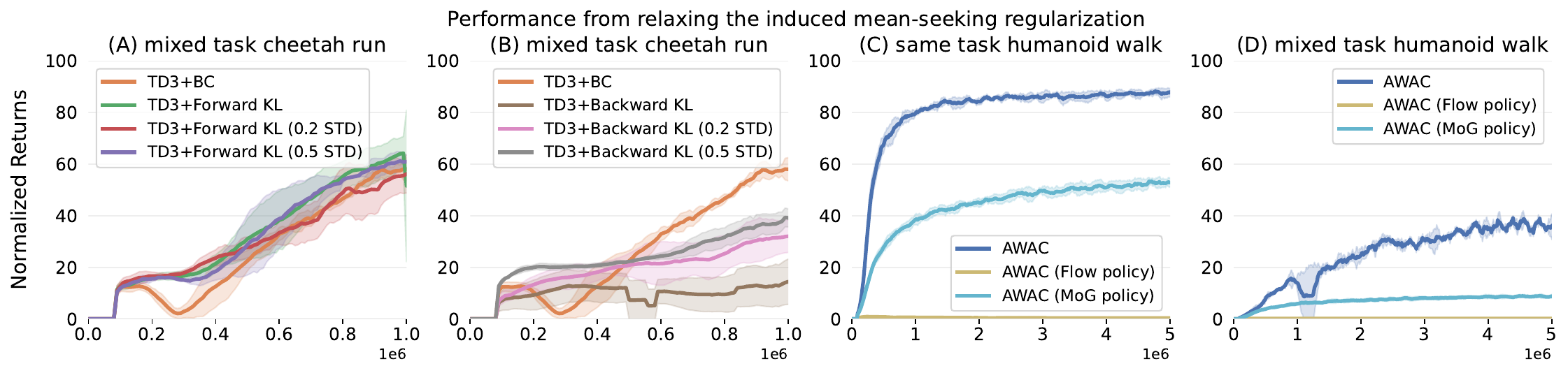}
    \caption{Results from using pre-trained density models (as described in Appendix~\ref{appD:pretrain_details}) to optimize for auxiliary policy improvement losses based on forward (A) and backward (B) KL divergences, and from increasing the policy's expressivity with affine Flows and Gaussian mixture distributions (C and D).  }
    \label{app:fig:kl_losses}
\end{figure}

One of our early hypotheses to what can cause over-conservativism was that Gaussian policies are a poor fit to multi-modal data distributions (e.g., as Fig.~\ref{sec3:fig:dist_evo_st_mt} clearly shows). Unfortunately, all our attempts to fix this issue, including the introduction of more expressive policy models (mixture of Gaussians, normalizing flows) or the usage of a reverse (mode-seeking) KL in the loss, led to performance collapse (Fig.~\ref{app:fig:kl_losses}). This failure was mostly due to early overfitting to the distribution matching objective while disregarding Q-function maximization, a phenomenon conceptually similar to posterior collapse in density modeling \citep{posterior_collapse_1, posterior_collapse_2}. This led us to discard such an hypothesis, as Gaussian policies seems to have a useful regularizing effect (at least early in training), and to focus only on ES as a solution to over-conservativism.

We focus on the role of mean-seeking and early attempts to overcome the action-averaging phenomenon illustrated in Figure~~\ref{sec3:fig:dist_evo_st_mt}. An extended version of such a figure with all action dimensions is shown in Figure~\ref{appE:fig:dist_evo_st_mt}. As detailed in the main text, the policy improvement optimization in TD3+BC, AWAC, and IQL can be be seen as performing a forward KL projection with respect to some inferred target distribution. Furthermore, they all parameterize a strictly unimodal policy, taking the form of either a Gaussian or a squashed Gaussian distribution. 
Hence, in case the target distribution displays significant multi-modality, using such models in conjunction with the mean-seeking nature of the forward KL loss would prevent ever closely matching part of the behavior data, leading to the displayed action-averaging phenomenon. Based on this considerations, we empirically analyze the effects of this induced mean-seeking regularization and the consequences from its relaxation. 


We find that, in spite of such problematic side-effects, mean-seeking regularization appears to be playing a crucial role to avoid premature convergence to an early suboptimal equilibrium. In particular, we analyze the effects of replacing the Gaussian policy and the forward KL objective with alternatives that allow to relax or overcome the mean-seeking regime. First, we analyze two simple variations of TD3 by adding to its original policy improvement objective from Equation \ref{eq:sec2:pi_pg_obj_cc} an auxiliary term to maximize either $\E_{a\sim \pi_\calB'(\cdot|s)}\left[\log \pi_\theta(a|s) \right]$ or $\E_{a'\sim \pi_\theta(\cdot|s)}\left[\log \pi_\calB'(a'|s) \right]$ using the pre-trained behavior model $\pi_\calB'$ described in Appendix~\ref{appD:pretrain_details}. We note that the first case 
is practically equivalent to the TD3+BC algorithm but with actions sampled from the learned behavior policy rather than the dataset, still falling in the mean-seeking regime. On the other hand, the auxiliary term in the second case corresponds to minimizing an \textit{inverse} KL with the behavior policy, $D_{KL}(\pi_\theta|\pi_\calB')$, and is akin to directly optimizing for the dual objective leading to AWAC's constrained advantage-maximizing targets $\pi_\calB^*$ \citep{awr_sem0}. 
In principle, this latter approach should fully preserve the canonical support constraint at the basis of most offline RL algorithms without the detrimental action-averaging, as it would allow the agent to focus on a single subset of $\pi^*_\calB$ right from the start.

As shown in part A of Figure~\ref{app:fig:kl_losses}, training with the mean-seeking forward KL objective expectedly yields very similar results to TD3+BC, validating the soundness of our formulated optimizations. However, when switching to the inverse KL objective in part B, we observe a severe degradation in performance, in direct contrast to its supposed theoretical benefits. We further analyze this phenomenon by pre-training increasingly mean-seeking proxies for $\pi_\calB$ by fixing (rather than learning) the standard deviation of the output mixture distributions of $\pi'_\calB$ to high values. The scope of these alternative parameterizations is to artificially emulate the mean-seeking regularization from behavior cloning and AWAC with the inverse KL objective, by forcing the pre-trained models themselves onto a mode-covering regime at the expense of accuracy. In particular, while the original $\pi'_\calB$ with a learned standard deviation achieves a log-likelihood of 1.09 bits per dimension, fixing the standard deviations to 0.2 and 0.5 only attain log likelihoods of 0.01 and -0.54 bits per dimension, respectively. However, as shown in parts A and B, using these worse models paradoxically leads to higher performance, suggesting that the induced mean-seeking regularization is actually \textit{an integral component} of current algorithms whose benefits far outweigh the downsides from modeling errors and action-averaging. As a validation check for our hypothesis, we also show that the performance of the forward KL TD3 variant, which already inherently encourages mean-seeking behavior regardless of the pre-trained models, does not improve and even slightly suffers from the resulting loss in precision. 
Finally, in parts C and D, we also analyze relaxing the mean-seeking regularization by increasing the expressivity of the policy's output distribution by parameterizing either a normalizing flow or a mixture of Gaussians. In particular, we consider either adding two additional affine flow layers conditioning on half the action dimensions and the state as in \citep{realnvp} or enlarging the output of the policy to represent parameters for five Gaussian heads. While these policies even outperform their traditional unimodal counterpart in the online setting, when used by offline algorithms such as AWAC, they seem to produce visibly slower learning with an analogous collapse in final performance to the one observed with the inverse KL objective, occurring even when training for the less diverse same-objective datasets.

\begin{figure}[t]
    \centering
    \includegraphics[width=0.95\linewidth]{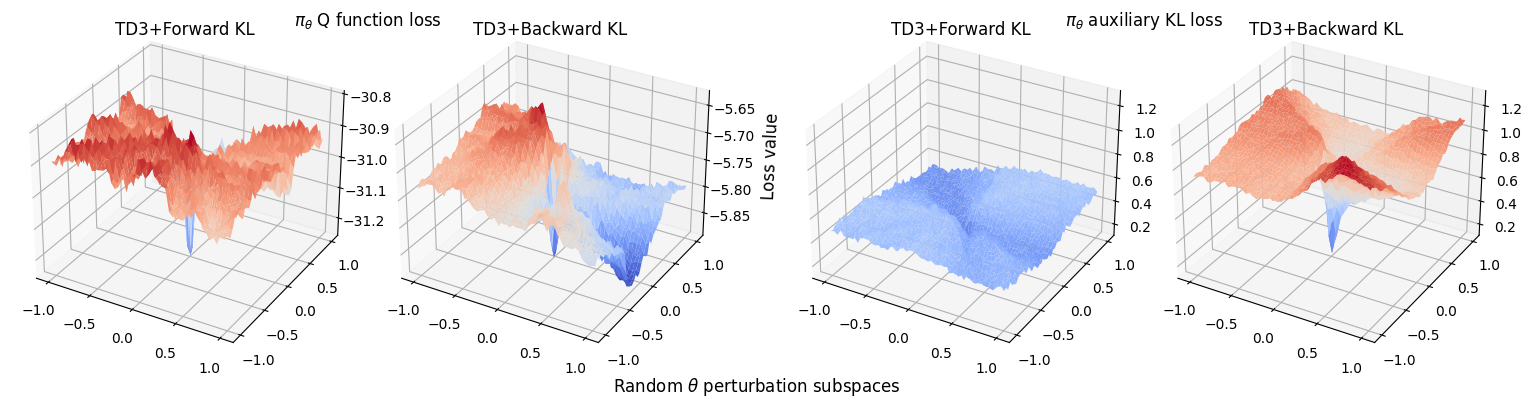}
    \caption{Two-dimensional loss surface projections using the visualization method from \citet{visualize_loss_land}. We produce visualizations for the Q function maximization and auxiliary KL loss for both TD3 modifications with auxiliary forward and backward KL losses}
    \label{sec3:fig:loss_surface}
\end{figure}%

This observed performance stagnation suggests that optimizing the offline policy with a tractable unregularized distribution matching objective leads to a phenomenon analogous to posterior collapse in density modeling \citep{posterior_collapse_1, posterior_collapse_2}. 
In particular, relaxing canonical constrains appears to enable the agent to initially focus on matching a likely suboptimal subset of the behavior policy, incurring the risk of converging to an early equilibrium. Instead, the mean-seeking objective induced by combining unimodal policies with the forward KL minimization avoids this early collapsing pull but seems to incur in the aforementioned unwarranted mode-averaging. This phenomenon can also be qualitatively identified following \citet{visualize_loss_land}, by visualizing the policy improvement loss surfaces induced by training with the TD3 variants employing the forward and backward KL auxiliary terms. As shown in Figure~\ref{sec3:fig:loss_surface}, the Q-function term of the actor loss attains a significantly lower absolute value and appears further from local convergence after training with the backward KL variant, reflecting its worse final performance. At the same time, the loss surface of the backward KL auxiliary term appears bound to a steep basin, in direct contrast with the much smoother minimum attained with its forward KL counterpart. Taken together, these visualizations appear to provide a further display of the implications of our hypotheses, corroborating how regularization-induced mean-seeking is an integral component of current algorithms rather than a flawed artifact. 

\subsection{Behavior Model Pre-training Details}

\label{appD:pretrain_details}
 \begin{table}[t]
\caption{Autoregressive density model hyper-parameters used to obtain a proxy for the unknown behavior policy.} 
\label{app:tab:hyper_pretrainedmod}
\vspace{10pt}
\begin{center}
\begin{tabular}{@{}lc@{}}
\toprule
\multicolumn{2}{c}{Density model hyper-parameters} \\ \midrule
batch size                      & $256$            \\
optimizer                       & Adam             \\
learning rate                   & $0.001$          \\
reserved validation data        & $15\%$           \\
maximum epochs                  & $100$            \\
encoder hidden layers           & $2$              \\
encoder hidden dimensionality   & $512$            \\
decoder hidden layers           & $2$              \\
decoder hidden dimensionality   & $64\times |A|$   \\
decoder mixture components      & $10$             \\
non-linearity                   & ReLU             \\ \bottomrule
\end{tabular}
\end{center}
\end{table}

 To produce Figure~\ref{sec3:fig:dist_evo_st_mt} in the main text and obtain the results in Appendix~\ref{appE:mean_seeking}, we pre-trained powerful autoregressive density models on the different \benchacro datasets to act as a proxy for $\pi_\calB$, which we denote $\pi'_\calB$. We employ a 85/15 split to partition the trajectories into the training and validation datasets and employ early stopping based on the epoch achieving the highest validation log-likelihood. Our model can be conceptually split into to components: i) an observation encoder, outputting a latent representation ii) an action decoder, outputting a distribution for each action dimension by conditioning on the output of the observation encoder and on all previous action dimensions. Hence, to sample any action, the decoder must be queried $|A|$ times in an autoregressive fashion. However, we still compute the density of any particular action in a single forward pass at training time by basing our architecture on the seminal MADE model from \citet{MADE}. The decoder output distribution we employ for each action dimension is a mixture of squashed Gaussians with ten independent components. Hence, for each action dimension, the autoregressive decoder outputs thirty values, representing the mean, log standard deviation, and weight logit for all mixture components. We found to obtain marginal gains with additional expressivity and that less-powerful models, such as simpler variational auto-encoders~\citep{vae} and affine Flows~\citep{realnvp}, are unable to closely fit the distribution of behavior policies in the mixed-objective datasets. We refer to the shared code and Table~\ref{app:tab:hyper_pretrainedmod} for further details and the employed hyper-parameters.


\end{document}